\documentclass[a4paper,fleqn]{cas-dc}
\usepackage[square]{natbib}

\usepackage{times}
\usepackage{soul}
\usepackage{url}
\usepackage[utf8]{inputenc}
\usepackage[small]{caption}
\usepackage{multirow}
\usepackage[section]{placeins}
\usepackage{bbding}
\usepackage{diagbox}
\usepackage{graphicx} 
\usepackage{adjustbox}
\usepackage{amsmath}
\usepackage{amsthm}
\usepackage {hyperref}
\usepackage{booktabs}
\usepackage{algorithm}
\usepackage{cas-common}
\usepackage{algorithmic}
\usepackage{float}
\usepackage{subfig}
\usepackage{color}
\usepackage{appendix}
\setcitestyle{numbers}
\usepackage{makecell}

\usepackage{amssymb}
\usepackage{indentfirst}
\usepackage[switch]{lineno}
\usepackage{psfrag}

\title[mode = title]{Tensor Decomposition Based Attention Module for Spiking Neural Networks}

\begin{document}




\author[1]{Haoyu Deng}
\ead{haoyu_deng@std.uestc.edu.cn}

\author[2]{Ruijie Zhu}
\ead{rzhu48@ucsc.edu}

\author[1]{Xuerui Qiu}
\ead{     sherry.qiu@std.uestc.edu.cn}

\author[1]{Yule Duan}
\ead{duanyll@std.uestc.edu.cn}

\author[1]{Malu Zhang}\fnref{cor}
\ead{maluzhang@uestc.edu.cn}

\author[1]{Liang-Jian Deng}\fnref{cor}
\ead{liangjian.deng@uestc.edu.cn}
\fntext[cor]{Corresponding authors: Malu Zhang, Liang-Jian Deng}
\address[1]{University of Electronic Science and Technology of China, 611731, China}
\address[2]{University of California, Santa Cruz, 95064, The United States}

\begin{abstract}[SUMMARY]
The attention mechanism has been proven to be an effective way to improve the performance of spiking neural networks (SNNs). However, from the perspective of tensor decomposition to examine the existing attention modules, we find that the rank of the attention maps generated by previous methods is fixed at 1, lacking the flexibility to adjust for specific tasks. To tackle this problem, we propose an attention module, namely Projected-full Attention (PFA), where the rank of the generated attention maps can be determined based on the characteristics of different tasks. Additionally, the parameter count of PFA grows linearly with the data scale. {PFA is composed of the \textit{linear projection of spike tensor} (LPST) module and \textit{attention map composing} (AMC) module.} In LPST, we start by compressing the original spike tensor into three projected tensors with learnable parameters for each dimension. Then, in AMC, we exploit the inverse procedure of the tensor decomposition process to combine the three tensors into the attention map using a so-called connecting factor. To validate the effectiveness of the proposed PFA module, we integrate it into the widely used VGG and ResNet architectures for classification tasks. Our method achieves state-of-the-art performance on both static and dynamic benchmark datasets, surpassing the existing SNN models with Transformer-based and CNN-based backbones. {Code for PFA is available at \href{https://github.com/RisingEntropy/PFA}{https://github.com/RisingEntropy/PFA} .}

\end{abstract}

\maketitle


\section{Introduction} 
Spiking neural networks (SNNs) are currently attracting the interest of academics due to their lower energy consumption and greater bio-interpretability compared to traditional artificial neural networks (ANNs)~ \cite{izhikevich2003simple, MAASS19971659}. The brain-inspired computation model gives it significant potential in processing temporal data. Though training a deep SNN is still a challenge, recent developments in SNNs introduce backpropagation  \cite{BP, Wu2019,zhang2021rectified,wei2023temporal,wu2021progressive,vtsnn} to relieve this issue to some degree. {Moreover, it makes the extending of ANN modules into SNNs possible, such as batch normalization and residual blocks.}

{As it is now feasible to incorporate ANN modules into SNNs while maintaining the inherent power efficiency, we can leverage these modules to augment the overall performance of SNNs.} A good practice for this is the introduction of residual architecture to SNNs. The well-known residual block proposed by  \citet{ResNet} makes it possible to train an ultra-deep ANN network. To introduce the residual block into the field of SNN, many variations \citep{SDRN,fang2021deep,ADRN} are presented from different perspectives, solving problems like gradient vanishing. {Moreover,} strategies in ANN, such as normalization \citep{DTBN} and architecture, could {also} effectively favor SNNs' performance, {thereby motivating} us to explore more useful techniques in ANNs for better-conducting SNNs.
\begin{figure}
    \centering
    \begin{minipage}{0.49\linewidth}
		\centerline{\includegraphics[scale=0.18]{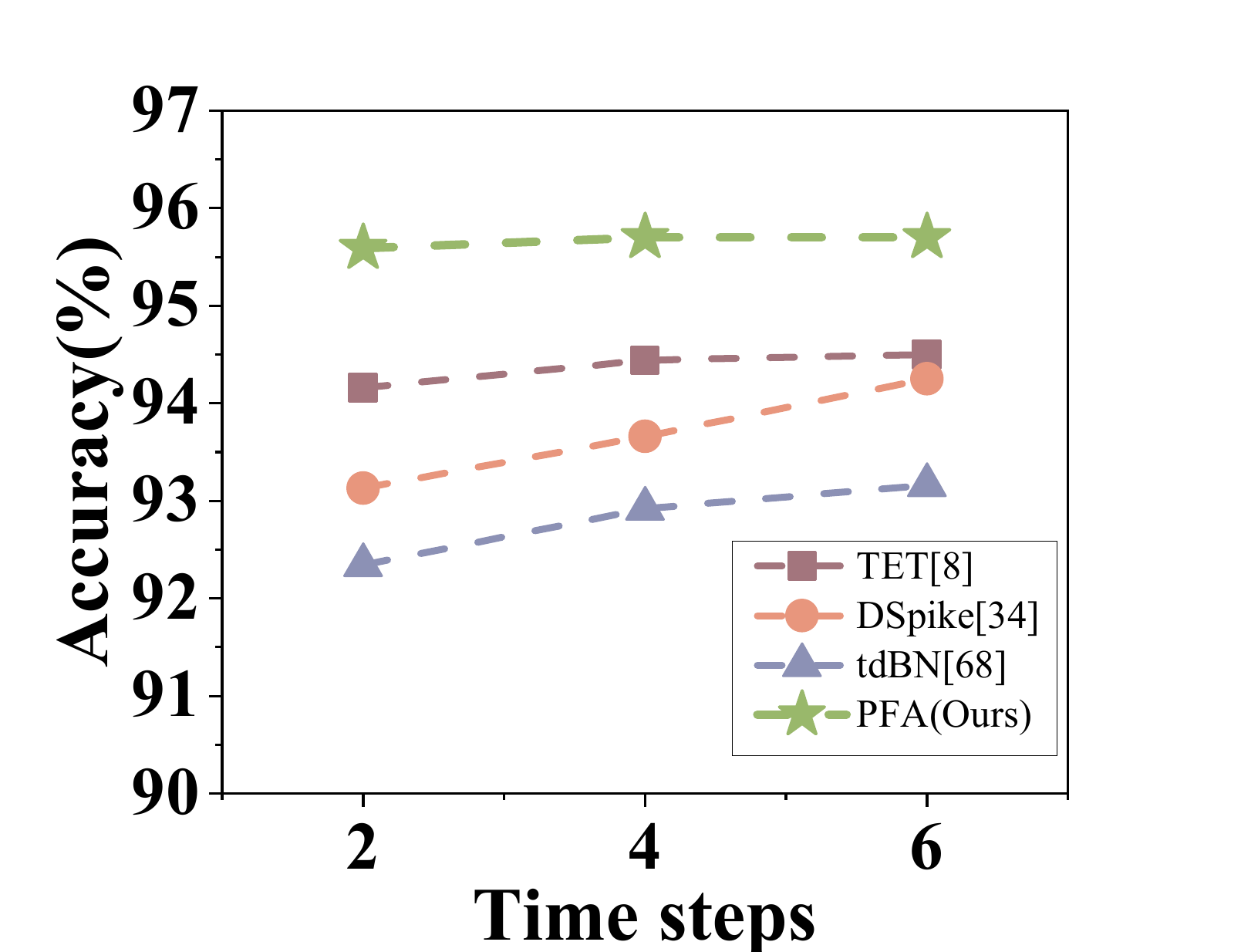}}
    \end{minipage}
    \begin{minipage}{0.49\linewidth}
		\centerline{\includegraphics[scale=0.18]{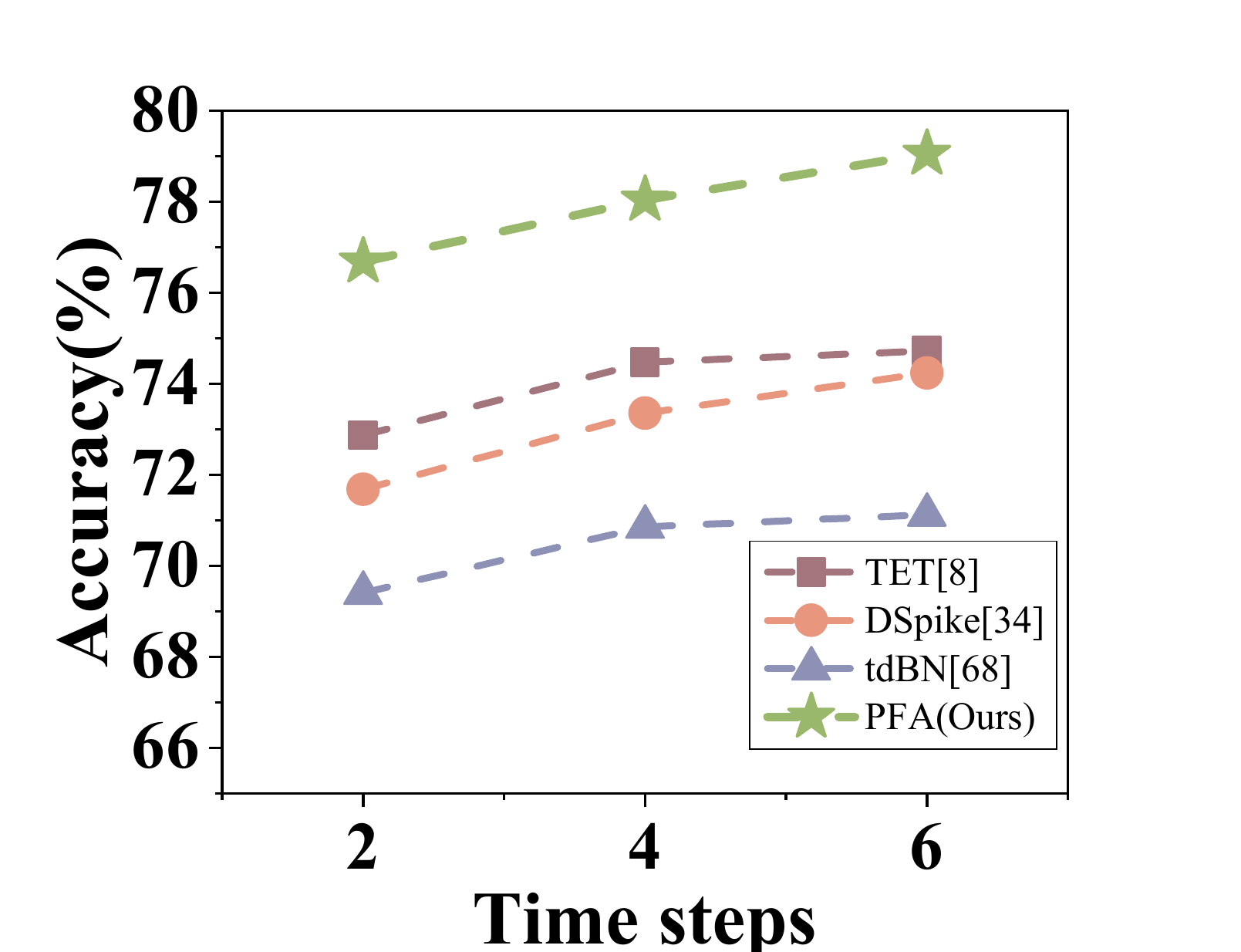}}
    \end{minipage}
    \caption{Accuracy on CIFAR10 (left) and CIFAR100 (right). Compared with other methods, PFA significantly improves network performance. }
    \label{fig:my_label}
\end{figure}
\par
{Apart from improving network architectures}, numerous other approaches can improve the performance of a network.  As one of the most promising techniques in ANNs, the attention mechanism can mimic the human ability to focus on certain things while ignoring others. { Three representative works \citep{TASNN,TCJA,TCS_Cai} have recently proven that the attention mechanism can successfully work in SNNs.} Especially,  \citet{TASNN} switch channel attention to the temporal dimension, revealing that paying attention to the temporal dimension is effective for performance enhancement.  \citet{TCJA} apply two 1D convolutions to a generated 2D tensor to achieve temporal-channel attention. Inspired by experimental observations on predictive attentional
remapping,  \citet{TCS_Cai} design the SCTFA module to assess the input tensor from temporal, channel, and spatial dimensions, achieving attention among these three dimensions. 
\par
Current software frameworks, {\emph{i.e.}, PyTorch and Tensorflow}, are boosted by GPU using parallel computation. {This makes it hard to realize event-driven SNNs} as events occur asynchronously. Based on these frameworks, as a trade-off, a popular approach is splitting the input event streams into slices and composing tensors for convenient later computation. This approach is adopted by all {attention-mechanism-related} works, showing a promising outcome. Since we are essentially dealing with tensor-formatted data flow, it is natural for us to consider involving tensor analysis theories to implement an attention module for SNNs, which is ignored by previous work. {By utilizing mathematical tools such as tensor decomposition theory, we can examine attention modules from a more abstract perspective, thereby identifying the shortcomings and limitations of previous methods.}


{
By expressing the composition of attention maps in previous methods through CANDECOMP/PARAFAC (CP) decomposition form, we found that they are essentially `rank-1' methods. When applied to various tasks, they lack a certain degree of flexibility. In this paper, we introduce a module called Projected-Full Attention (PFA), which is capable of generating attention maps with a rank other than 1. Specifically, PFA consists of two components: Linear Projection of Spike Tensor (LPST) and Attention Map Composing (AMC). LPST is responsible for generating $R$ sets of projections from the input tensor using a small number of parameters. AMC is responsible for creating attention maps using these $R$ sets of projections, where $R$ is called The connecting factor representing the rank concept of the CP decomposition and is a hyper-parameter that can be adjusted based on the specific task.}

{The key contributions of this paper are outlined as follows:}
\begin{itemize}
    \item We propose PFA, a module that can achieve temporal-channel-spatial attention. In contrast to previous approaches, the ranks of the attention maps generated by PFA are not fixed at 1; instead, the rank can be chosen based on the specific task. Furthermore, the parameter count of PFA grows linearly with the data scale, and the computational cost is equivalent to that of a single standard convolution layer.
    \item We present a comprehensive theoretical analysis, focusing on two critical aspects: the rank of tensors and $R$, the connecting factor. These analyses lead to a tailored selection criterion, providing effective insights into experimental outcomes and offering {a} valuable guidance for future applications.
    \item  We conduct thorough experiments on both static and dynamic datasets. The results demonstrate the {effectiveness of PFA which achieves} state-of-the-art (SOTA) accuracy on both dynamic and static datasets. Ablation studies further validate the structural reasonableness of PFA. Additionally, we visualize the attention map generated by PFA to intuitively illustrate attention distribution across temporal, channel, and spatial dimensions.
\end{itemize}

\section{Related Works}
\textbf{Spiking Neural Network:} 
Spiking Neural Networks (SNNs), the third generation of neural networks  \cite{izhikevich2003simple, MAASS19971659}, offer a closer emulation of the human brain's efficiency by utilizing discrete spikes for information transmission \citep{Zhan2023}. This unique method of communication allows for enhanced {energy-saving feature} compared to traditional Artificial Neural Networks (ANNs), which rely on continuous signals. The temporal representation capabilities of SNNs have been a focal point in research \citep{TASNN, deng2021temporal, Zhan2023}, drawing from seminal neurodynamic models such as the Hodgkin-Huxley (H-H) model  \cite{HH} and further developed in works like Izhikevich  \cite{izhikevich2003simple} and Leaky Integrate and Fire (LIF)  \cite{LIF}.
These models underscore SNNs' potential in capturing time-dependent patterns in data, a feature not inherently present in ANNs. The challenges in training SNNs, due to their non-differentiable spiking behavior, have also been addressed in recent literature, with gradient-based optimization methods  \cite{STBP,wu2019direct,Wenjie2024} and ANN-to-SNN conversion techniques  \cite{Conversion1, Conversion2,wang2022signed,wu2021tandem} being key developments. In addition to research on training methods, there are many other studies dedicated to improving the performance of SNNs, such as attention mechanisms.

\textbf{Attention Mechanism in SNNs:}
Since Google proposed recurrent models of visual attention in 2014  \cite{Google_Attention_2014}, the attention mechanism become a potent tool to increase the performance of a neural network, which gives distinct parts of the input data varied weights. For instance, the squeeze-and-excitation network (SENet)  \cite{SENet} {gives different weights to different channels of input data and significantly improves the performance.} From this fact, it is of vital significance to introduce the attention mechanism into SNNs. \citeauthor{gated} apply the attention mechanism to the encoding layers of SNNs \cite{gated}. To achieve temporal-wise attention, inspired by SENet, TA-SNN  \cite{TASNN} is proposed. Besides, TCJA  \cite{TCJA} uses two 1D convolutions to pay attention to both channel dimension and temporal dimension to prolong the attention mechanism. Nevertheless, they fail to consider three crucial aspects of SNNs, \emph{i.e.} temporal, channel, and spatial. By changing SENet and leaving affection in the temporal direction,  \citet{TCS_Cai} and  \citet{yao2023attention} successfully apply temporal-channel-spatial attention to SNNs. {While the methods of predecessors have been very successful, they do not take into account the fact that the input data is a high-order tensor. }
Examining data from a tensor perspective can provide a more mathematically abstract viewpoint, revealing additional properties. As described in Section \ref{sec:motivation}, the methods employed by previous researchers result in `rank-1' attention maps, lacking the flexibility and specificity for different tasks.

\noindent\textbf{Tensor Decomposition:} Tensor decomposition methods  \citep{8918128, 7891546, kolda} have undergone years of development and achieved remarkable success. {Many prior studies  \citep{Kossaifi2020, Yang2016DeepMR, Novikov2015TensorizingNN, kossaifi2017, babiloni2023factorized,liu2023tensor,wang2023inertial} have embraced these techniques to optimize neural network architectures or uncover versatile modules with multiple functionalities. } \citeauthor{Kossaifi2019FactorizedHC}, for instance, conducted a comprehensive review of convolution kernels from the perspective of tensor decomposition and introduced an innovative convolution module that can seamlessly extend to higher dimensions. {\citeauthor{Lau2024} propose a new decomposition method to enable the direct use of the depth-wise convolutional layer with large kernels in the attention module, without requiring any extra blocks \citep{Lau2024}.
Apart from using tensor decomposition methods to improve network structures, tensor decomposition theory can also impose certain constraints on the data itself \citep{NEURIPS2022_acbfe708, xu2023,chenwanli}. \citeauthor{xu2023} leverage tensor decomposition to predict real-time traffic flow, demonstrating its effectiveness in handling complex patterns \cite{xu2023}. \citeauthor{chenwanli} employ tensor CP decomposition to generate an attention map for segmentation, achieving significant improvements \cite{chenwanli}. These methods either use tensor decomposition to reduce the parameter count in processing high-dimensional data or leverage low-rank properties to enhance performance.} In this paper, drawing inspiration from these prior endeavors, we leverage tensor decomposition methods to enhance the field of SNNs. We design a lightweight attention module, designed to be both parameter-efficient and computationally efficient.

\section{Motivation and Method}
\subsection{Motivation\label{sec:motivation}}
{Tensor CP decomposition \citep{CPdeconstruction}, is a mathematical technique employed to decompose a high-order tensor into a sum of rank-one tensors. Mathematically, for a tensor denoted as $\mathcal{X}\in \mathbb{R}^{D_1,D_2,\dots ,D_n}$, the CP decomposition can be expressed as:
\begin{equation}
    \mathcal{X} \approx \sum_{r=1}^{R} U_{1,r}^{(D_1)}\circ U_{2,r}^{(D_2)}\circ\dots \circ U_{n,r}^{(D_n)}.
    \label{eq:cp}
\end{equation}
In this equation, $U_{1,r}^{(D_1)}\circ U_{2,r}^{(D_2)}\circ\dots \circ U_{n,r}^{(D_n)}$ represents the $r$-th rank-one tensor, where the symbol $\circ$ signifies the outer product of vectors, and $R$ denotes the rank. Specifically, $U_{n,r}^{(D_1)}$ represents a vector corresponding to the $n$-th dimension in the $r$-th rank-one tensor, with a dimensionality of $D_1$.}
\par
{
Rethinking previous attention modules \citep{TASNN,TCJA,TCS_Cai} from the perspective of CP decomposition, we can find that they are all rank-one methods. For the sake of clarity, {we denote an all-one vector with a dimensionality of $D$ as $I^{(D)}$. }Specifically, for TA-SNN  \citep{TASNN}, the temporal attention module for SNNs yields the attention map $\mathcal{A}_{\text{TA-SNN}}$ for the input tensor $\mathcal{X}$ as follows:
\begin{equation}
\mathcal{A}_{\text{TA-SNN}} = I^{(HW)}_{S}\circ I^{(C)}\circ\mathcal{F}_{T}^{(T)}(\mathcal{X}).
\label{eq:tasnn}
\end{equation}
Here, $\mathcal{F}_{T}^{(T)}(\mathcal{X})$ represents a function that generates the temporal attention vector with a dimensionality of $T$. The same mathematical form can also be used to reconfigure the TCJA \citep{TCJA} and SCTFA \citep{TCS_Cai} method. For TCJA, the attention map $\mathcal{A}_{\text{TCJA}}$ can be written as:
\begin{equation}
    \mathcal{A}_{\text{TCJA}} = I^{(HW)}_S\circ\mathcal{F}^{(C)}_C(\mathcal{X})\circ\mathcal{F}^{(T)}_T(\mathcal{X}).
    \label{eq:tcja}
\end{equation}
For SCTFA, the attention map $\mathcal{A}_{\text{SCTFA}}$ is:
\begin{equation}
    \mathcal{A}_{\text{SCTFA}} = \mathcal{F}^{(HW)}_S(\mathcal{X})\circ\mathcal{F}^{(C)}_C(\mathcal{X})\circ\mathcal{F}^{(T)}_T(\mathcal{X}).
    \label{eq:sctfa}
\end{equation}
}
\par
{
By comparing Equation \ref{eq:cp} and Equation \ref{eq:tasnn}, \ref{eq:tcja}, \ref{eq:sctfa}, we can easily see that the $R$ of Equation \ref{eq:tasnn}, \ref{eq:tcja}, \ref{eq:sctfa} is fixed at 1, making them special cases of Equation \ref{eq:cp}. This motivates us to design an attention module that the rank of the attention map is not fixed at 1, \textit{allowing us to control the attention effect by adjusting the rank of {the} attention map{;} namely, we wish our proposed module {could} be written in a form like:}
\begin{equation}
    \mathcal{A}_{PFA} = \sum_{r=1}^{R}U^{(HW)}_{s,r}\circ U^{(C)}_{c,r}\circ U^{(T)}_{t,r}.
\end{equation}
{Through} this way, we can use $R$ to control certain properties of the attention map, such as low rank and the degree of compression of the original tensor. From the perspective of tensor decomposition, previous methods can be seen as special cases of our method when $R=1$. However, due to the absence of summation terms in previous methods, adjustments tailored to the specificity of the dataset cannot be made, lacking a certain degree of flexibility. The flexibility mentioned is crucial, as demonstrated by the experiments presented in Section \ref{sec:dis_on_r}. The results indicate that the optimal choice for $R$ varies across different datasets.}

\subsection{Projected Full-Attention (PFA)}
\begin{figure}
    \centering
    \includegraphics[scale=0.25]{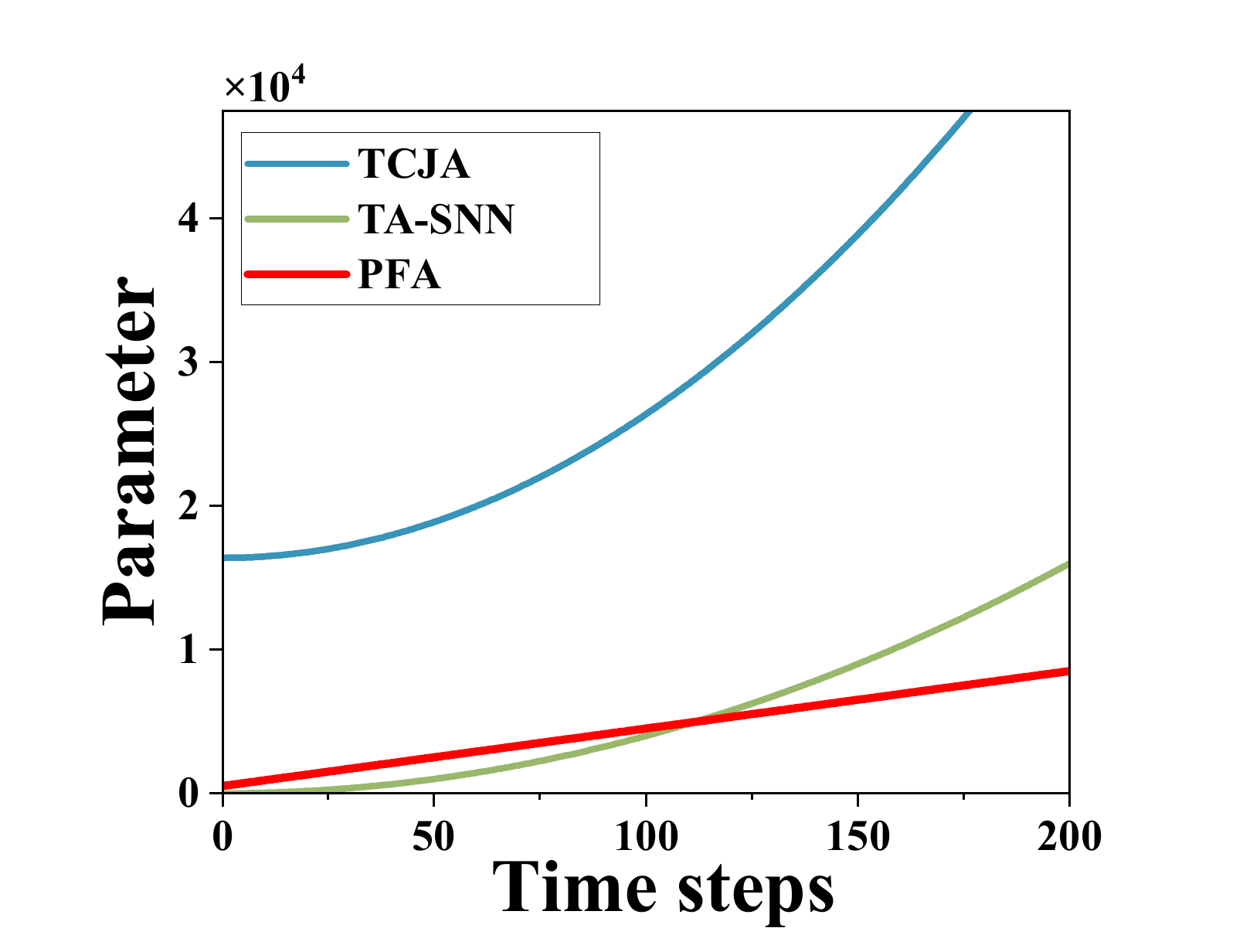}
    \caption{A simple comparison curve of parameter quantity growth among TCJA  \cite{TCJA}, TA-SNN  \cite{TASNN}, and PFA. The parameter scale of PFA increases linearly. In this figure, the channel number is fixed at 128.}
    \label{fig:para}
\end{figure}
\begin{figure*}
    \centering
    \includegraphics[scale=0.08]{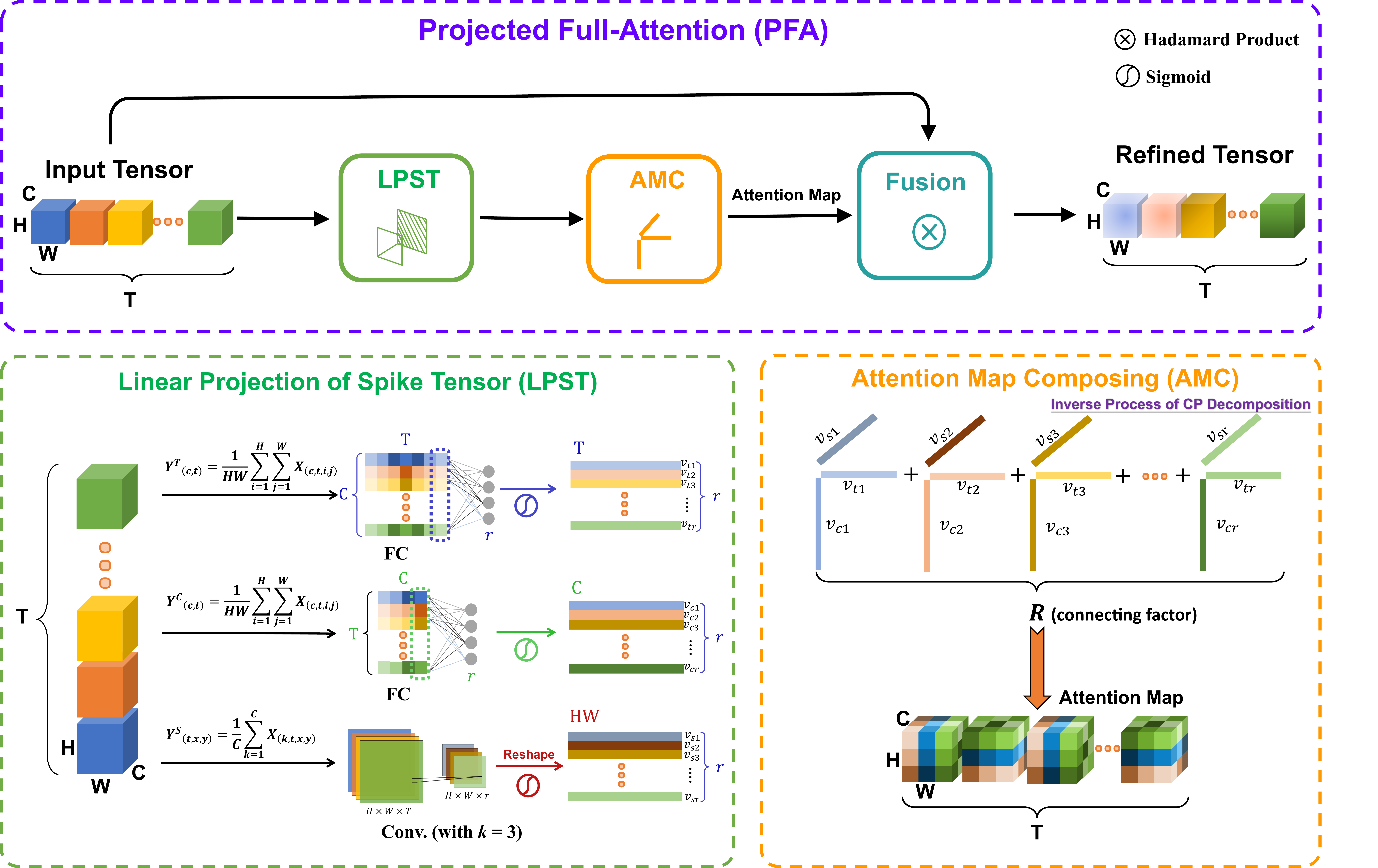}
    \caption{The detailed workflow of PFA. The input tensor is first sent to Linear Projection of Spike Tensor (LPST) {module} to generate three projections and split the three projections into corresponding vectors. In Attention Map Composing (AMC) {module}, these vectors are composed into the final attention map through the inverse process of CP decomposition. The attention is fused with the input tensor to obtain the refined tensor by Hadamard product. }
    \label{fig:main}
    \vspace{-0.5cm}
\end{figure*}

{In this section, we will present an overview of our proposed PFA module and detailed descriptions of two sub-modules of PFA.} PFA is composed {of} two sub-modules, Linear Projection of Spike Tensor (LPST) and Attention Map Composing (AMC). LPST, as its name suggests, produces projections from input tensor $\mathcal{X}$ for AMC in the form of matrices. {The AMC module splits the matrices from LPST to vectors to compose the final attention map.}
\par
\begin{figure}
    \centering
    \includegraphics[width=0.9\linewidth]{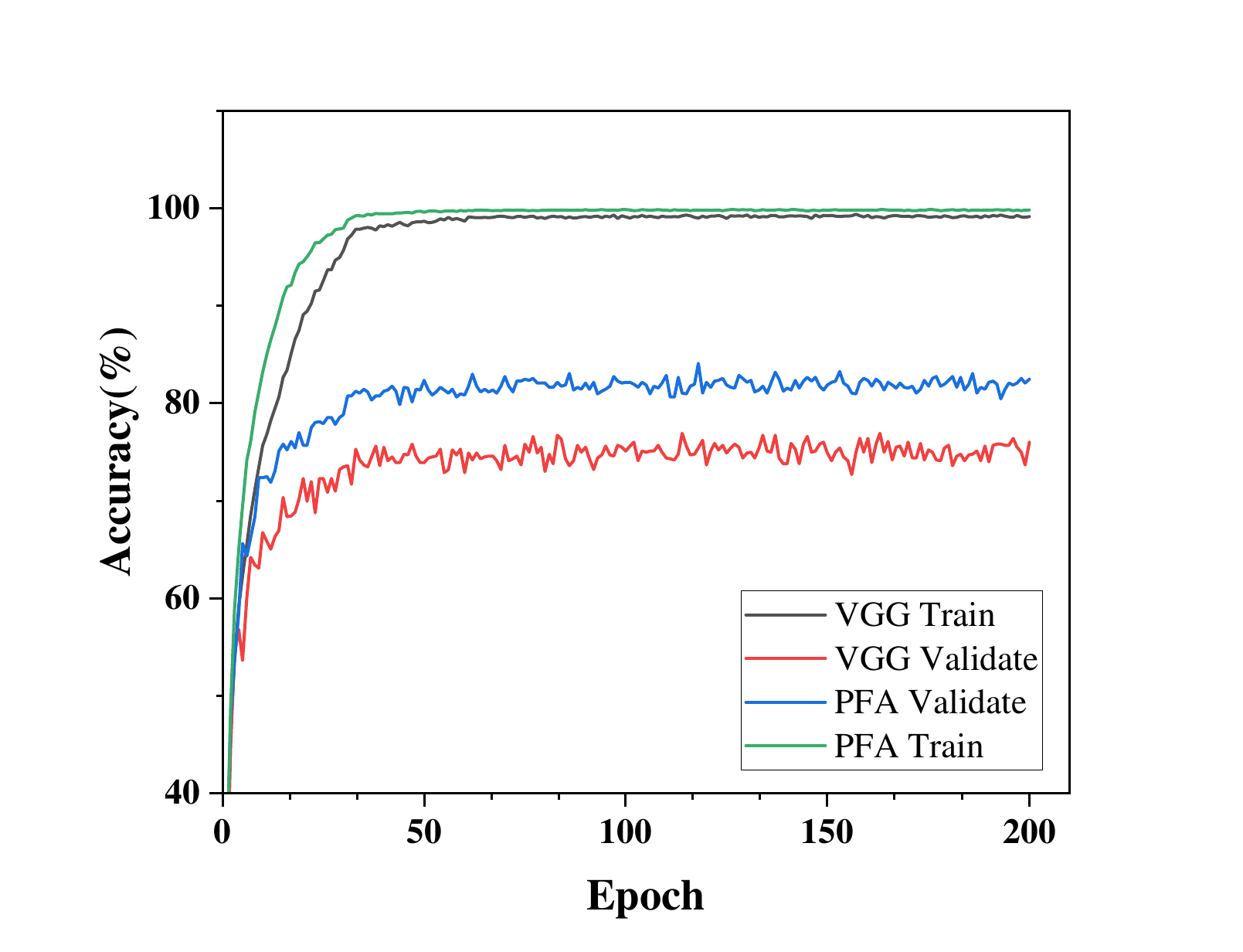}
    \caption{{Comparison of accuracy between the vanilla VGG network and VGG network with PFA modules on training and {validation set}s. While the training set accuracy remains consistently high for both models, the {validation set} accuracy shows a significant improvement with the addition of PFA modules, suggesting its efficacy in mitigating overfitting.}}
    \label{fig:overfitting}
    \vspace{-0.5cm}
\end{figure}
The core of the attention mechanism is to give essential parts of the input data higher weights while ignoring less-related parts to some degree. This is particularly important for SNNs, especially for data collected by DVS cameras. {Data from DVS camera often contains} more noise, which greatly affects network performance. Thanks to the feature of CP-decomposition, PFA can effectively suppress noise. {Also}, as shown in Figure \ref{fig:overfitting}, the inexact representation of {the} input tensor can help reduce overfitting, which is very severe in SNNs. In our proposed PFA, we first design the LPST module to project the higher-order tensor (generally 3D) to three lower-order tensors, \emph{i.e.}, 2D matrices. {Using} learnable parameters, LPST can selectively reserve essential information while ignoring irrelevant ones. In AMC, we split the matrices into $R$ (called connecting factor here) vectors. Three vectors from three matrices of temporal, channel, and spatial dimensions respectively are used to conduct the outer product to compose a part of the final attention map. The final attention map is the sum of these $R$ parts. We fuse this attention map with the origin input to obtain the refined tensor. 
\par

The workflow of PFA is plotted in Figure \ref{fig:main}. In what follows, we will give the details of LPST and AMC.
\subsubsection{Linear Projection of Spike Tensor (LPST)}
Tensor decomposition is a vital approach to represent a tensor and its latent features. {Therefore,} it is natural for us to utilize {tensor-decomposition-related} techniques to explore the data characteristics. A typical technique is the CP decomposition. Whatever way we adopt, it is essential to generate a `projection' to extract significant components of the input tensor to yield an attention map later.          

In this part, we need to construct three linear projections of the input data stream along spatial, channel, and temporal dimensions. In LPST, we apply two fully connected (FC) layers and a convolution layer to project the input tensor for simplicity. Note that here we simply use single-layer FCs and a convolution layer.  What we do here is just a simple projection with learnable parameters rather than complex function fitting. {In this way, we retain the required information and reduce the number of parameters.}

For the temporal dimension, we first squeeze the input tensor ($H\times W\times C\times T$) in the following way:
\begin{equation}
    Y^T_{(c,t)}=\frac{1}{HW}\sum\limits_{i=1}^{H}\sum\limits_{j=1}^{W}\mathcal{X}_{(c,t,i,j)},
\end{equation}
where $Y^T_{(c,t)}$ denotes the output matrix, $H$ and $W$ denote the spatial sizes, $\mathcal{X}_{(i,j,c,t)}$ denotes the input tensor. After the squeeze operation, the tensor is converted to a matrix in the shape of $C\times T$. Then, we apply an FC layer to reserve $R$ necessary information in each column. A sigmoid activation function is applied after the FC layer. Finally, we obtain a projection matrix in the shape of $r\times T$.


For the channel dimension, the operation is similar to that of the temporal dimension. We squeeze the input tensor by:
\begin{equation}
    Y^C_{(c,t)}=\frac{1}{HW}\sum\limits_{i=1}^{H}\sum\limits_{j=1}^{W}\mathcal{X}_{(c,t,i,j)},
\end{equation}
where $Y^C_{(c,t)}$ denotes the output matrix, $C$ denotes the number of channel. Then, we pass the $T\times C$ matrix through an FC layer to acquire the projected matrix in the shape of $r\times T$ with the sigmoid activation function applied.

For spatial dimension, the channel is squeezed by:
\begin{equation}
    Y^S_{(t,x,y)}=\frac{1}{C}\sum\limits_{k=1}^{C}\mathcal{X}_{(k,t,x,y)}.
\end{equation}
Afterward, we employ a convolution to the squeezed tensor, treating $T$ as the input channel and $R$ as the output channel. Then, we reshape the output tensor from the convolution into a matrix in the shape of $HW\times r$. The sigmoid activation function is applied. 
\subsubsection{Attention Map Composing (AMC)}
Once we obtain the three projections corresponding to three dimensions, a necessary step is to compose an attention map using the three projections. Here, we adopt the reverse process of the tensor CP decomposition.
In the CP decomposition, a tensor $\mathcal{X}\in \mathcal{R}^{HW\times C\times T}$ can be represented as:
\begin{equation}
    \mathcal{A}_{s,c,t} = \sum_{r=1}^{R}U^{(HW)}_{s,r}\circ U^{(C)}_{c,r}\circ U^{(T)}_{t,r},
\end{equation}
where $U_{si}$, $U_{ci}$ and $U_{ti}$ denote the $i$-th vectors in the spatial, channel, and temporal projection tensors respectively, $\mathcal{A}$ is the attention map, and $\circ$ is the outer product of vectors. {In particular, $R$, the connecting factor, can be viewed as the rank concept in the tensor CP decomposition (this process is plotted in Figure \ref{fig:main}).} It is worth noting that the tensor CP decomposition is just an approximate representation of tensors, which perfectly meets our demands. \textit{{$R$ can control the accuracy of our representation of the input tensor.} That is, the larger $R$ is, the more accurate the representation is, and vice versa, which perfectly realizes the attention mechanism we need.} In the experiment part, we discuss the choice of connecting factor $R$ and its effect in detail.
\subsection{Parameter and Computational Cost Analysis}
This section presents an analysis of the parameter and computational cost of PFA. PFA has an advantage in terms of parameter quantity. For an input tensor with the shape of $H\times W\times C\times T$, the overall parameter amount is: 
\begin{equation}
    \underbrace{C\times R}_{\text{MLP of }Y^T}+\underbrace{T\times R}_{\text{MLP of }Y^C}+\underbrace{k^2\times T\times R}_{\text{Conv of }Y^S},
\end{equation}
where $k$ is the convolution kernel size. Note that $R$ is limited to a small fixed value according to our theoretical analysis in Section \ref{sec:analysis}, and the parameter size grows linearly with $T$ and $C$. Figure \ref{fig:para} shows the parameter growth curve of PFA and compares several other techniques. 
\par
{PFA is also lightweight in computational burden. The time complexity of PFA can be calculated in the following way:
\begin{equation}
        \underbrace{3HWTC}_{\text{Obtain } Y^T,Y^C,Y^S}+\underbrace{2TCR}_{\text{MLP of } Y^T,Y^C}+
        \underbrace{HWk^2TR}_{\text{Conv of } Y^S }+\underbrace{RHWTC}_{\text{AMC}}.
\end{equation}
It is worth noting that a standard convolution operation typically entails a computational burden of $HWk^2TC_{in}C_{out}$. In this context, given that $R$ is not large, a PFA module incurs a computational load that is smaller than that of a convolution operation since $C$ is usually in the range of several hundred.}
\par
\subsection{Theoretical Analysis on $R$\label{sec:analysis}}
\begin{figure}
    \centering
    \includegraphics[scale=0.25]{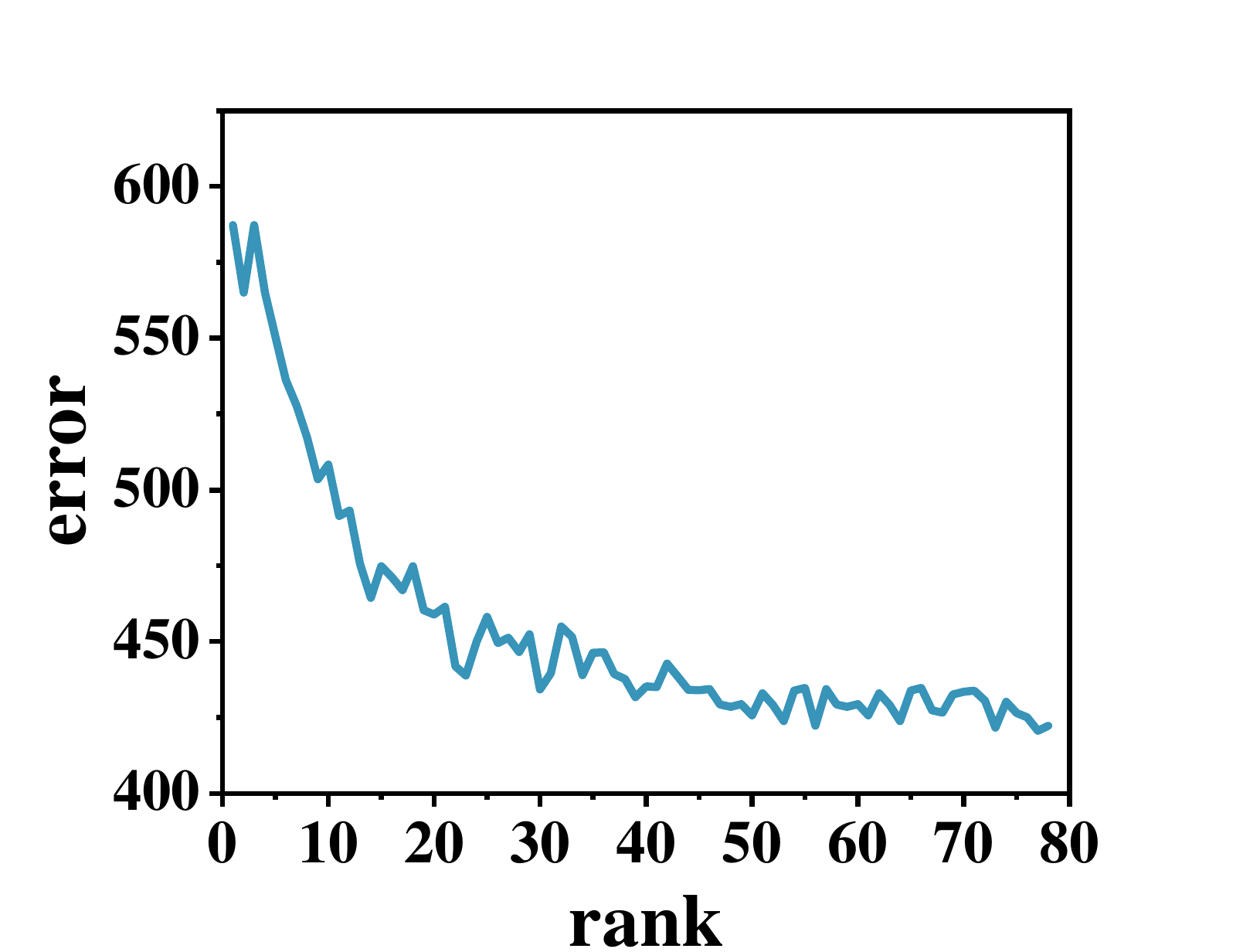}
    \caption{A toy example of the effect of different ranks on the approximation outcome. \textit{error} is measured under $\ell_2$ norm. The norm starts to cover when the rank exceeds 30.}
    \label{fig:input_tensor_rank}
\end{figure}
In light of the fact that CP decomposition provides only an approximate representation of a tensor, and the quality of this approximation relies on the chosen factor, denoted as $R$, it is essential to establish a clear guiding principle for determining the optimal value of $R$. This section delves into our approach for gaining insights into the characteristics of PFA and streamlining its future applications.

Our exploration begins by adopting the methodology put forth by  \citet{scalable} to investigate the rank of the input data. However, due to the recognized NP-hard nature of computing the CP decomposition of a tensor  \cite{np_hard1,CPdeconstruction}, a direct calculation from the tensor becomes impractical. {To surmount this challenge, we employ a gradient descent technique to approximate the original tensor by systematically experimenting with different rank values, ultimately selecting the one that yields the best approximation for the tensor.}
To provide further clarity, consider an element of the tensor represented as:
{\begin{equation}
    x_{ijk}\approx \sum\limits_{r=1}^{R}a_{i,r}b_{j,r}c_{k,r}.
\end{equation}
We treat this as an optimization problem, with $a_{i,r}$, $b_{j,r}$, and $c_{k,r}$ considered as parameters. We then optimize the loss function defined as:
\begin{equation}
    \mathcal{L}=\frac{1}{2}\sum_{(i,j,k)\in \Omega}(x_{i,j,k}-\sum\limits_{r=1}^{R}a_{i,r}b_{j,r}c_{k,r})^2.
\end{equation} 
To perform optimization, we compute the first-order partial derivative of the loss with respect to a parameter (e.g., $a_{i,r}$), given by:
\begin{equation}
    \frac{\partial \mathcal{L}}{\partial a_{i,r}}=-\frac{1}{2}\cdot 2\cdot (x_{i,j,k}-\sum\limits_{r=1}^{R}a_{i,r}b_{j,r}c_{k,r})\cdot b_{j,r}c_{k,r}.
\end{equation}
Finally, we apply the gradient descent method for optimization:
\begin{equation}
    a_{i,r}'\leftarrow a_{i,r}-\mu\cdot \frac{\partial \mathcal{L}}{\partial a_{i,r}},
\end{equation}}
{
Here, $\mu$ represents the step size used in the gradient descent process. This approach allows us to effectively estimate the rank of the tensor while circumventing the computational complexities associated with CP decomposition.
}

We take a piece of data from the dataset CIFAR10DVS as a toy example and discover its rank. In this case, we adopt $\mu=0.0001$ and iterate the process 1000 times to get the final optimized parameters. We utilize $\ell_2$ norm as the metric to judge the approximation effect. As shown in Figure \ref{fig:input_tensor_rank}, the error reduces as the rank increases, and starts to converge when the rank exceeds 30. This toy example shows that the rank of the input tensor is around 30, indicating that we need a rank exceeding 30 {to represent it
using CP decomposition precisely. It should be noted that accurately representing the original tensor for the attention mechanism is not conducive to improving network performance. 
} 
 The core idea of the attention mechanism is to focus on the important parts while ignoring the secondaries. Representing raw tensors too precisely is counterproductive. On the one hand, overly precise characterization of input tensors is detrimental to neglecting unimportant parts. On the other hand, it may lead to severe overfitting. As shown in Figure \ref{fig:overfitting}, the vanilla VGG network exhibits severe overfitting, achieving nearly 100\% accuracy on the training set but less than 80\% on the {validation set}. However, upon incorporating the PFA modules, there is not a substantial improvement in training set accuracy. Nevertheless, there is a significant increase in {validation set} accuracy. This observation underscores the benefit of utilizing less precise tensor representations with PFA {to mitigate overfitting}. Therefore, a good choice for $R$ should not be too large. This is a very useful conclusion. On the one hand, it tells us the range of finding $R$. On the other hand, it also ensures that PFA can obtain good results without excessive calculation. 
\par

As shown in our experiment (Figure \ref{fig:rank}), though the rank of the attention map differs on dynamic datasets like CIFAR10DVS and static datasets like CIFAR100, the best option for $R$ is not very big. Further observing the experimental results, we find that when the value of $R$ exceeds a certain threshold (approximately $\frac{T}{2}$, $T$ is the time step), the larger $R$ is, the worse the effect will be. The above analysis of the experimental results proves the correctness of our conclusion. Considering the above analysis, we propose a principle for selecting $R$: for dynamic datasets, the search for the optimal value of $R$ is performed within the range of no more than $T$. Specifically, we focus on identifying the best $R$ value around $\frac{T}{2}$, considering the varying content across different time steps. In the case of static datasets, the search for the optimal $R$ value begins from 1 and extends until the best value is determined. Since static datasets involve duplicated frames, this search aims to find the most suitable $R$ value for the low-rank condition.
\section{Experiment}
We evaluate the classification performance of our PFA on two static datasets: CIFAR10  \cite{krizhevsky2009learning}, CIFAR100  \cite{krizhevsky2009learning} and two dynamic datasets: CIFAR10DVS  \cite{CIFAR10DVS}, NCaltech-101  \cite{NCALTECH101}. 
Details of the datasets, network architecture, data augmentation, loss function, and pre-process
procedure is introduced in this section. Extensive experiment results are also presented.
\begin{table}[htbp]    
    \centering
    
    \begin{tabular}{cccc}
    \toprule[2pt]
         Dataset&Learning Rate&Epoch&Batch Size \\
         \midrule
        CIFAR10 &0.1&200&128\\
         
          CIFAR100 &0.1&200&64\\
          CIFAR10DVS &0.0001&200&32\\
          NCaltech-101 &0.0001&200&32\\
         

         \bottomrule[2pt]
    \end{tabular}
    \caption{Hyper-parameter settings of PFA. 
    }
    \label{tab:conf}
\end{table}
\subsection{Datasets and Training Details}
\subsubsection{Datasets}
We have conducted experiments on both {static and dynamic datasets} for object classification. We train and {validate} PFA on a workstation equipped with one RTX 3090. For different datasets, the hyper-parameters are listed in Table \ref{tab:conf}. The summaries of datasets and augmentation involved in the experiment are listed below.

\noindent\textbf{CIFAR 10/100} consist of 50k training images and 10k testing images with the size of $32 \times 32 $ \cite{krizhevsky2009learning}. We use ResNet-19 for both CIFAR10 and CIFAR100. Random horizontal flips and crops are applied to the training images for augmentation. Moreover, cutout is also used for  augmentation which is the same as  \cite{deng2021temporal}\\
\noindent\textbf{CIFAR10-DVS} converts 10,000 frame-based images of 1010 classes into event streams with the dynamic vision sensor. Since the CIFAR10DVS dataset  \cite{CIFAR10DVS} does not divide training and testing sets, we split the dataset into 9k training images and 1k
test images and reduced the spatial resolution from $128 \times 128$ to $48 \times 48$  \cite{deng2021temporal}.
We use VGGSNN for neuromorphic datasets CIFAR10DVS. 
In each frame,  horizontal flipping and mixup are adopted, where the probability of Flipping is set to 0.5. Then, we randomly select one augmentation among rolling, rotation, cutout, and shear, where the random rolling range is 5 pixels, and the degree of Rotation is sampled from the uniform distribution, which is the same as the Ref  \cite{deng2021temporal}. \\
\noindent\textbf{NCALTECH-101} is also converted from the original version of Caltech-101  \cite{NCALTECH101} with a slight change in object classes to avoid confusion. The NCaltech-101 consists of 100 object classes plus one background class. We apply the 9: 1 train-validation split as CIFAR10DVS. {The VGGSNN is} applied to neuromorphic datasets NCALTECH-101.  Moreover, the augmentation is identical to CIFAR10DVS.
{\noindent\textbf{Fashion-MNIST} is a more demanding successor of the famous MNIST dataset. Fashion-MNIST contains 70,000 static grayscale images of 10 different categories of clothing, each image being of size 28x28 pixels. We utilize the simple example network from the Spikingjelly framework, which comprises two convolution layers and a fully-connected layer, to evaluate PFA on this dataset.}
\subsubsection{Loss function}
We used the cross-entropy loss for the neuromorphic datasets CIFAR10DVS and NCALTECH-101. In order to get a better performance in static datasets (CIFAR 10 / 100), we used a Temporal Efficient Training  \cite{deng2021temporal} loss function. It can be concluded as follows:
\begin{equation}
    \mathcal{L}_{\mathrm{TET}}=\frac{1-\lambda}{T}\cdot  \sum_{t=1}^T \mathcal{L}_{\mathrm{CE}}[\boldsymbol{O}(t), \boldsymbol{y}] +\frac{\lambda}{T}\cdot \sum_{t=1}^T \mathcal{L}_{\mathrm{MSE}}(\mathbf{O}(t), \phi),
\end{equation}
where  $T$ is the total time steps, $\mathcal{L}_{\mathrm{CE}}$ and $\mathcal{L}_{\mathrm{MSE}}$ denote the cross-entropy loss and mean square error. And $\boldsymbol{y}$ represents the target
label. Moreover, $\mathbf{O}(t)$ and $\phi$ are the output and a constant used to regularize the membrane potential distribution. Additionally, we set $\phi = V_{th}$
in our experiments. In practice, we use a hyper-parameter $\lambda$ to adjust the proportion of the regular term.

\begin{figure}[h]
    \centering
    \begin{minipage}{0.49\linewidth}
		\centerline{\includegraphics[scale=0.13]{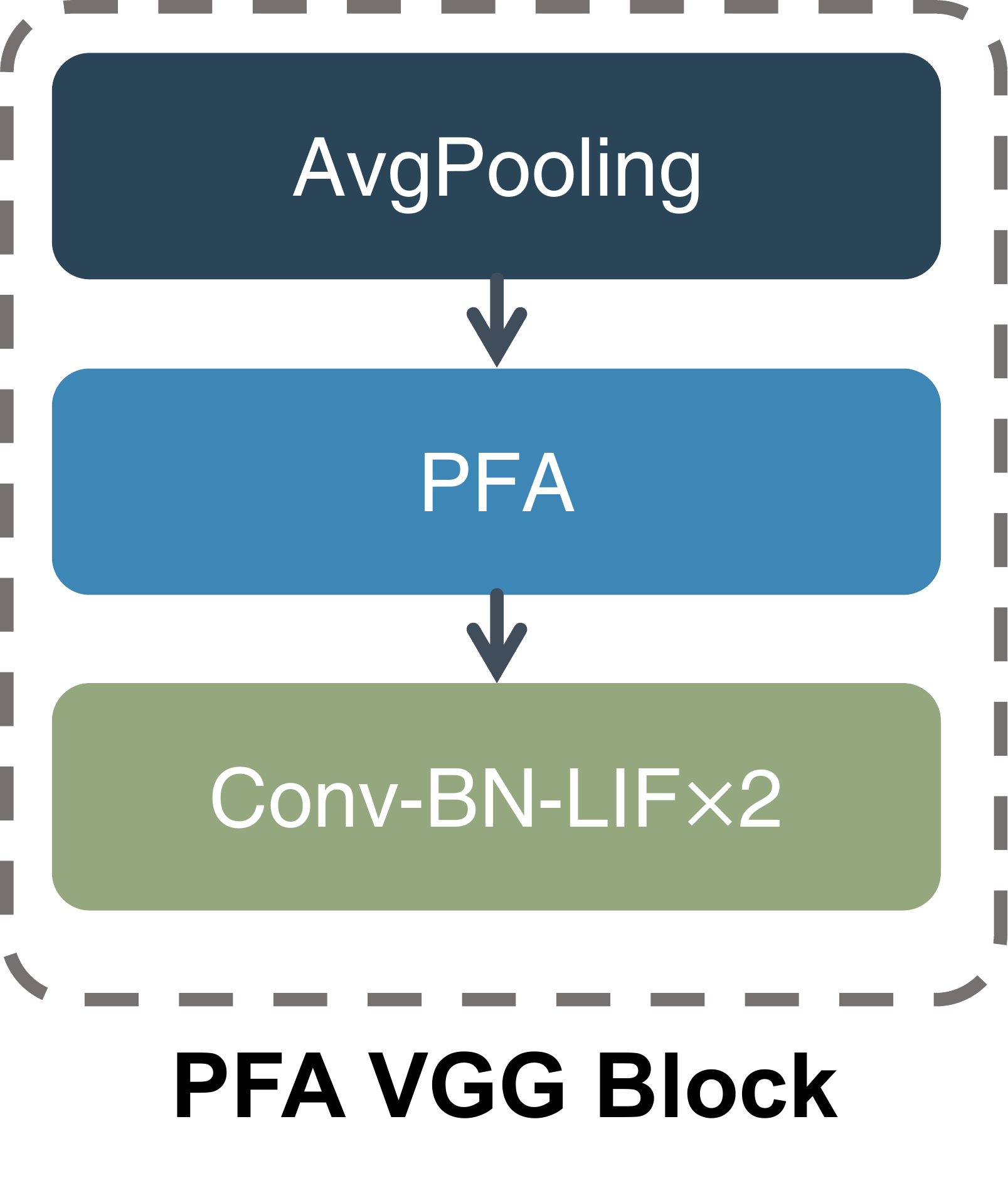}}
    \end{minipage}
    \begin{minipage}{0.49\linewidth}
		\centerline{\includegraphics[scale=0.13]{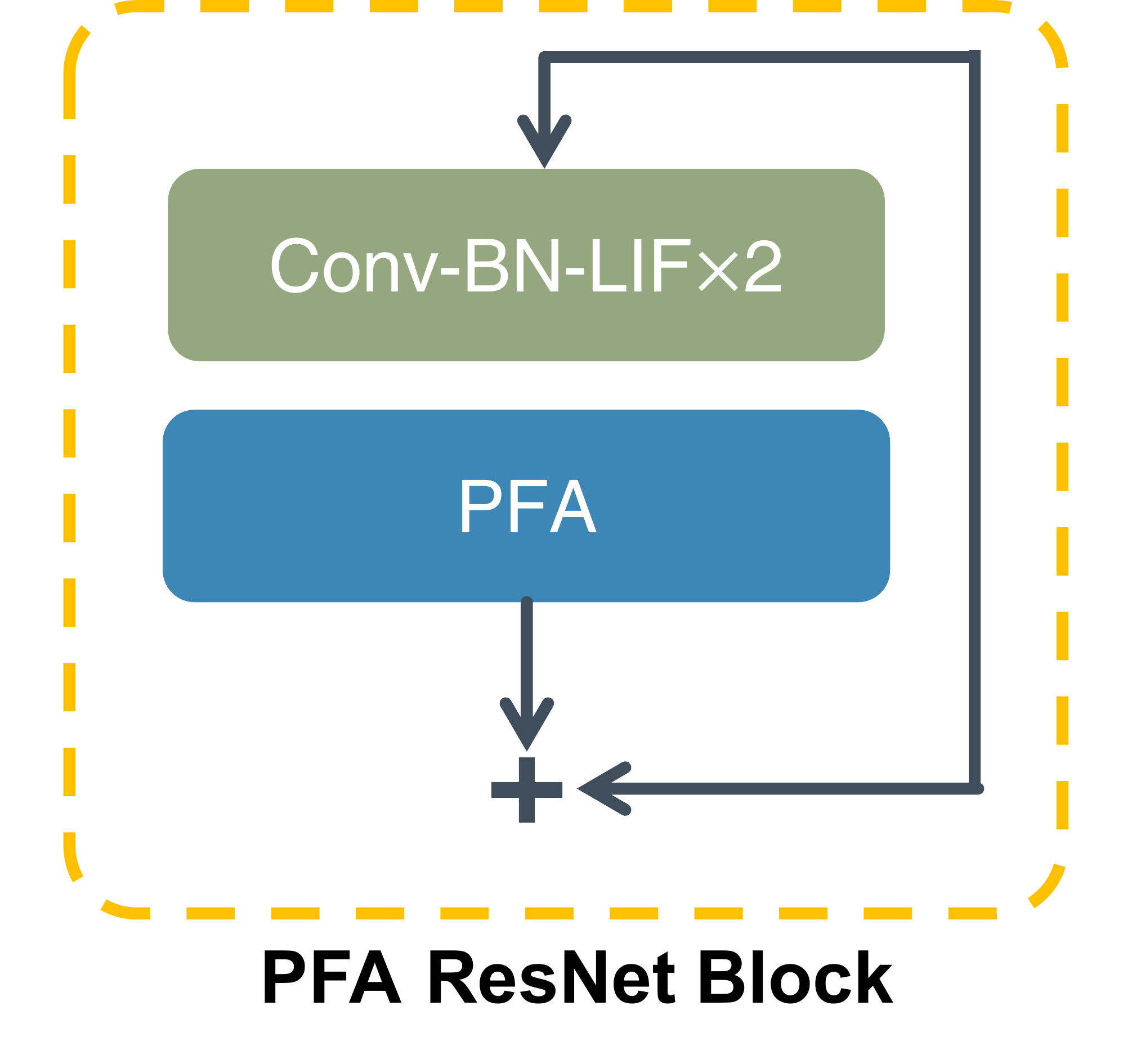}}
    \end{minipage}

    \caption{Schematic diagram of the network structure. On the left is the convolution layer of the VGG network, and on the right is the convolution layer of Resnet. }
    \label{fig:arctecture}

\end{figure}

\subsubsection{Network Architectures}
To evaluate PFA, we integrate it into VGG-type  \cite{VGG} and ResNet  \cite{ResNet} architectures, which are previously utilized in  \cite{fang2021incorporating,TCJA,deng2021temporal,yao2023attention}. Figure \ref{fig:arctecture} presents the basic blocks involved in VGG and ResNet architectures. Five PFA modules are inserted into a VGG-type network after each average pooling operation. In {the} ResNet architecture, PFA is inserted before the residual connection in the last two layers. When splitting the event flow data into frames, we simply integrate the input in the time domain. Particularly, our work is based on the software framework SpikingJelly  \cite{SpikingJelly}. Taking notice that frames in CIFAR10-DVS and NCaltech-101 are entirely occupied by their content, we utilized a pooling layer as the first layer of the network to reduce memory consumption.
\subsection{Comparison with Existing SOTA Works}
\begin{table*}[htbp]
\centering

   \begin{adjustbox}{max width=\linewidth} 
\begin{tabular}{clcccc}
\toprule[2pt] 
Dataset                    & Method                     & Type                              & Architecture & Time steps            & Accuracy 
\\
\midrule
\multirow{11}{*}{CIFAR10/CIFAR100}
& SIDP~ \cite{rathienabling}\textsuperscript{\textit{ICLR-2019}}                        & Hybrid training                           & ResNet-20/VGGSNN                      & 250/125                & 95.4/76.4                   \\
& Diet  \cite{rathi2021diet}\textsuperscript{\textit{TNNLS-2021}}                & Spike-based   BP                           & ResNet-20                        & 10/5                & 92.5/64.0                    \\
 & tdBN   \cite{zheng2021going}\textsuperscript{\textit{AAAI-2021}}                       & Spike-based BP                    & ResNet-19                & 6                    & 93.1/71.1                       \\

&Dspike  \cite{li2021differentiable}\textsuperscript{\textit{NeurIPS-2021}}       & Spike-based   BP & ResNet-18       & 6                    & 94.2/74.2                     \\
 &NAS-SNN   \cite{kim2022neural}\textsuperscript{\textit{ECCV-2022}}                       & Spike-based BP                    & NAS                & 5                    & 92.7/73.0                      \\   
                          
& TET  \cite{deng2021temporal}\textsuperscript{\textit{ICLR-2022}}    & Spike-based   BP & ResNet-19      & 6                    & 94.5/74.7 \\
& DSR  \cite{meng2022training}\textsuperscript{\textit{CVPR-2022}}    & Spike-based   BP & ResNet-19      & 20                    & 95.4/78.5 \\
& Spikformer  \cite{zhou2023spikformer}\textsuperscript{\textit{ICLR-2023}}    & Spike-based   BP & Spiking-ViT      & 4                    & 95.5/78.2 \\
                           \cmidrule{2-6}
& \multirow{3}{*}{\textbf{PFA}}      & \multirow{3}{*}{\textbf{Spike-based   BP}} & \multirow{3}{*}{\textbf{ResNet-19}}       
& \textbf{6} &\textbf{95.7/79.1}    \\
&                            &                                   &       & \textbf{4}                    &\textbf{95.7/78.1}       \\
&                            &                                   &       & \textbf{2}                    &\textbf{95.6/76.7}       \\
                           \midrule
\multirow{9}{*}{CIFAR10DVS} 
& LIAF-Net~ \cite{wu2021liaf}\textsuperscript{\textit{TNNLS-2021}} & Spike-based BP                           & LIAF-Net   & 10               & 70.4                     \\
& tdBN   \cite{zheng2021going}\textsuperscript{\textit{AAAI-2021}}                      & Spike-based BP                    & ResNet-19                & 10                   & 67.8                      \\
&TA-SNN~ \cite{TASNN}\textsuperscript{\textit{ICCV-2021}}                      & Spike-based BP                    & LIAF-Net               & 10                  & 72.0                     \\

& Dspike  \cite{li2021differentiable}\textsuperscript{\textit{NeurIPS-2021}} & Spike-based BP                    & VGGSNN                       & 10                   & 75.4               \\
& TET  \cite{deng2021temporal}\textsuperscript{\textit{ICLR-2022}}                       & Spike-based BP                    & VGGSNN                  & 10                   & 83.2                    \\
& DSR~ \cite{meng2022training}\textsuperscript{\textit{CVPR-2022}}                & Spike-based BP                    & VGGSNN                     & 10                    & 77.3                    \\
& TCJA~ \cite{TCJA}\textsuperscript{\textit{TNNLS-2024}}             & Spike-based BP                    & VGGSNN                 & 10                    & 80.7                 \\
& Spikformer  \cite{zhou2023spikformer}\textsuperscript{\textit{ICLR-2023}}    & Spike-based   BP & Spiking-ViT      & 16                    & 80.9 \\
                            \cmidrule{2-6}
& \textbf{PFA}                       & \textbf{Spike-based BP}                    & \textbf{VGGSNN}                        & \textbf{14}                    & \textbf{84.0}                     \\

 \midrule
\multirow{3}{*}{NCaltech-101} 
& SALT~ \cite{kim2021optimizing}\textsuperscript{\textit{Neural Netw-2021}} & Spike-based BP                           & VGGSNN    & 20               & 55.0                    \\
& TCJA~ \cite{TCJA}\textsuperscript{\textit{TNNLS-2024}}                     & Spike-based BP                    & VGGSNN               & 14                   & 78.5                     \\

                            \cmidrule{2-6}
& \textbf{PFA}                       & \textbf{Spike-based BP}                    & \textbf{VGGSNN}                        & \textbf{14}                    & \textbf{80.5}                     \\
\bottomrule[2pt]       
\end{tabular}
\end{adjustbox}

\caption{Compare with existing works. Our method improves network performance across all tasks.}
\label{tabresult}
\end{table*}

{To validate the effectiveness of our PFA, we conducted a comparative analysis with several state-of-the-art SNNs, including those employing hybrid training and spike-based backpropagation. The results for both static image data and dynamic data classification tasks are summarized in Table \ref{tabresult}. Specifically, the results for the static image dataset are presented at time steps 2, 4, and 6.}
\par
For CIFAR10 and CIFAR100 datasets, our PFA outperforms prior works in terms of accuracy. Notably, PFA achieves superior performance on CIFAR10 and CIFAR100, with respective improvements of 1.2\% and 4.4\% over the TET method  \cite{deng2021temporal}{ which utilizes} the same network architecture and time steps. Furthermore, PFA surpasses the DSR method  \cite{meng2022training} while requiring only 10$\times$ fewer time steps. Additionally, our attention-based SNN demonstrates better performance than the Spiking-ViT backbone  \cite{zhou2023spikformer}.
\par
\begin{table}
    \centering
    \resizebox{\linewidth}{!}{
    \begin{tabular}{lcc}
    \toprule[2pt]
        Method&Time Steps&Accuracy\\
        \midrule
         ST-RSBP\citep{Spike-Train}$^{\text{NIPS-2019}}$&  400 &90.1\\
         LISNN\citep{LISNN}$^{\text{IJCAI-2020}}$&20&92.1\\
         PLIF\citep{fang2021incorporating}$^{\text{ICCV-2021}}$&8&94.4\\
         TCJA\citep{TCJA}$^{\text{TNNLS-2024}}$&8&\textbf{94.8}\\
        \midrule
        \textbf{PFA (Ours)}&8&94.5\\
        \bottomrule[2pt]
    \end{tabular}
    }
    \caption{{Accuracy on Fashion-MNIST\citep{FashionMNISTAN} dataset.}}
    \label{tab:fashion}
    \vspace{-0.5cm}
\end{table}
{
For the Fashion-MNIST dataset, as depicted in Table \ref{tab:fashion}, while the accuracy of our proposed PFA falls slightly short of TCJA \citep{TCJA}, it still surpasses previous methodologies utilizing longer time steps \citep{Spike-Train, LISNN} or incorporating learnable spike thresholds \citep{fang2021incorporating}. We hypothesize that the relatively diminutive image size of the Fashion-MNIST dataset curtails the efficacy of spatial attention. This conjecture aligns with the findings from our ablation experiments, indicating that spatial attention plays a pivotal role in determining PFA performance.
}
\par
For the dynamic dataset CIFAR10DVS, our method outperforms previous approaches using binary spikes by 3.1\%, even with fewer time steps. For the NCaltech101 dataset, we achieved an impressive 80.5\% top-1 accuracy, far surpassing the performance of prior work  \cite{TCJA}.
\par
To further illustrate the advantages of our PFA, {we compare it} with previous works at various time steps. As depicted in Figure \ref{fig:my_label}, we compare PFA with Spike-based BP SNNs. Notably, our PFA exhibits a significant advantage at shorter simulation time steps, owing to its superior representation capability.

\par
{We observe from the perspective of simulated time steps that PFA can achieve satisfactory performance even with a smaller number of simulated time steps.} For the CIFAR-10 dataset, {simply} utilizing 2 simulated time steps with PFA can surpass the performance of previous works, including transformer-like structures such as Spikeformer. On the CIFAR-100 dataset, compared to the top-performing Spikeformer, {PFA lags by} only 0.1\% accuracy with 4 time steps. On DVS datasets like CIFAR-10DVS and NCaltech, employing 14 simulated time steps, PFA significantly outperforms previous methods. On CIFAR10-DVS, PFA surpasses Spikeformer by 3.1\%, and compared to the TET \citep{deng2021temporal} method with a similar structure, it achieves a higher accuracy by 0.8\%. On the NCaltech-101 dataset, PFA outperforms the previous best TCJA method by 2\%.

\par
In summary, PFA offers substantial advantages over previous methods. When applied to attention modules like TCJA  \citep{TCJA}, PFA significantly enhances resistance to overfitting, leading to improved accuracy on the {validation set}. {Although Spikeformer, which is based on the Spike ViT architecture, performs better than many CNN-based techniques, an especially interesting breakthrough is the incorporation of PFA into traditional CNN architectures like VGG, which can even outperform Spikeformer's \cite{zhou2023spikformer} performance.

\subsection{{Experiments on Image Generation Tasks}}
\begin{figure}
    \centering
    \includegraphics[width=\linewidth]{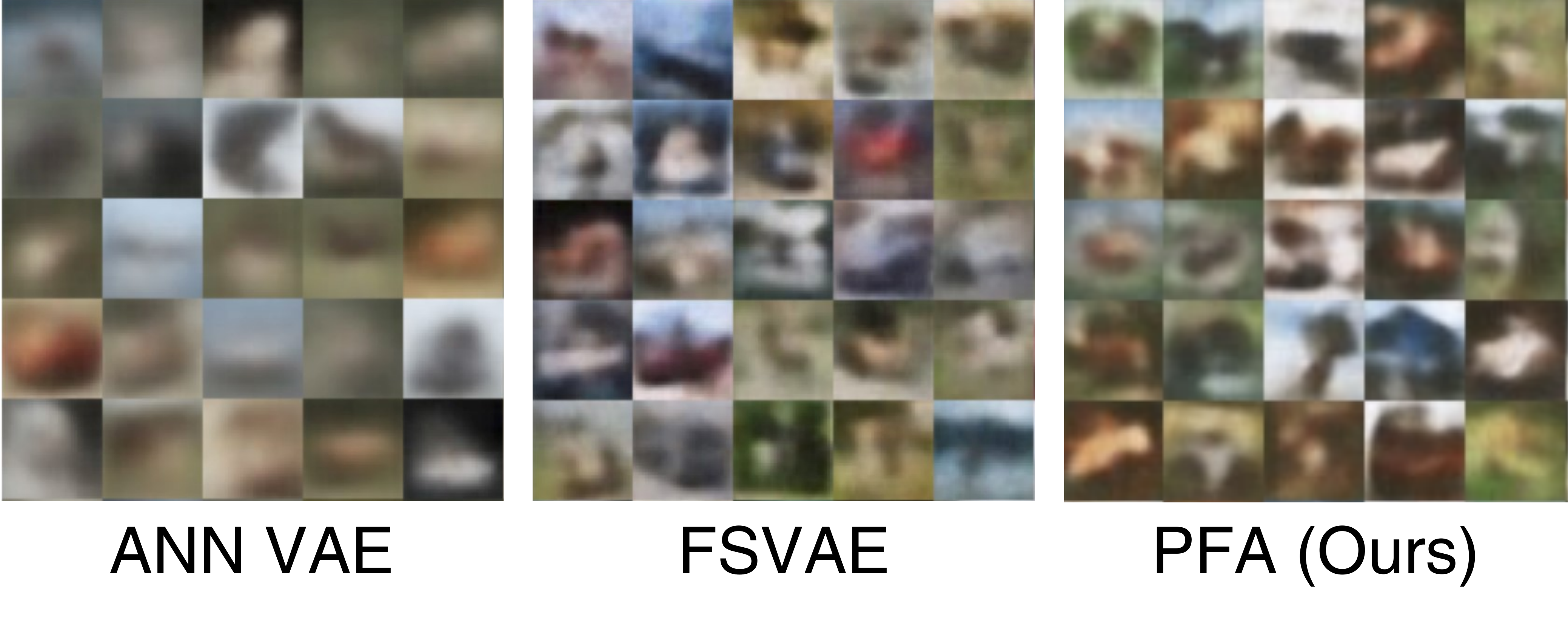}
    \caption{{Generated images of ANN VAE, FSVAE\citep{fsvae}, and our PFA on CIFAR10 \citep{cifar10} dataset.}}
    \label{fig:fsvae_img}
    \vspace{-0.5cm}
\end{figure}
\begin{table}[h]
    \centering
    \begin{tabular}{ccccc}
    \toprule[2pt]
        {Dataset}&{Method}&IS$\uparrow$&FID$\downarrow$&FAD$\downarrow$\\
        \midrule
        \multirow{4}{*}{MNIST\citep{minist}}&ANN\citep{fsvae}\textsuperscript{AAAI-2022}&5.95&112.5&17.09\\
        &FSVAE\citep{fsvae}\textsuperscript{AAAI-2022}&6.21&97.06&35.54\\
        &TCJA\citep{TCJA}\textsuperscript{TNNLS-2024}&6.45&100.8&19.39\\
        &\textbf{PFA (Ours)}&\textbf{6.97}&\textbf{96.82}&\textbf{15.12}\\
        \midrule
        \multirow{4}{*}{\makecell{Fashion-\\MNIST\citep{FashionMNISTAN}}}&ANN\citep{fsvae}\textsuperscript{AAAI-2022}&4.58&123.7&18.08\\
        &FSVAE\citep{fsvae}\textsuperscript{AAAI-2022}&4.55&\textbf{90.12}&15.75\\
        &TCJA\citep{TCJA}\textsuperscript{TNNLS-2024}&\textbf{5.61}&93.41&12.46\\
        &\textbf{PFA (Ours)}&5.35&97.3&\textbf{11.97}\\
        \midrule
        \multirow{4}{*}{CIFAR10\citep{cifar10}}&ANN\citep{fsvae}\textsuperscript{AAAI-2022}&2.59&229.6&196.9\\
        &FSVAE\citep{fsvae}\textsuperscript{AAAI-2022}&2.94&175.5&133.9\\
        &TCJA\citep{TCJA}\textsuperscript{TNNLS-2024}&3.73&170.1&100.4\\
        &\textbf{PFA (Ours)}&\textbf{3.84}&\textbf{166.4}&\textbf{92.83}\\
    \bottomrule[2pt]
    \end{tabular}
    \caption{{Comparison with existing methods on image generation tasks.}}
    \label{tab:fsvae_tab}
    \vspace{-0.5cm}
\end{table}
{To further validate the efficacy of PFA, we conduct image generation tasks on MNIST \citep{minist}, Fashion-MNIST \citep{FashionMNISTAN}, and CIFAR10 \citep{cifar10} using fully spike variation autoencoders \citep{fsvae} (FSVAE) . We integrated our proposed PFA as the first layer of FSVAE while maintaining other settings identical to the original network. The experimental results are outlined in Table \ref{tab:fsvae_tab}. Notably, PFA exhibits superior performance metrics on both MNIST \citep{minist} and CIFAR10 \citep{cifar10} datasets. Additionally, we visually depict the generated images in Figure \ref{fig:fsvae_img}, further corroborating the enhancements facilitated by PFA.}
\subsection{Discussion on $R$\label{sec:dis_on_r}}
\begin{figure}[h]
    \centering
        \begin{minipage}{0.48\linewidth}
        \centerline{\includegraphics[scale=0.19]{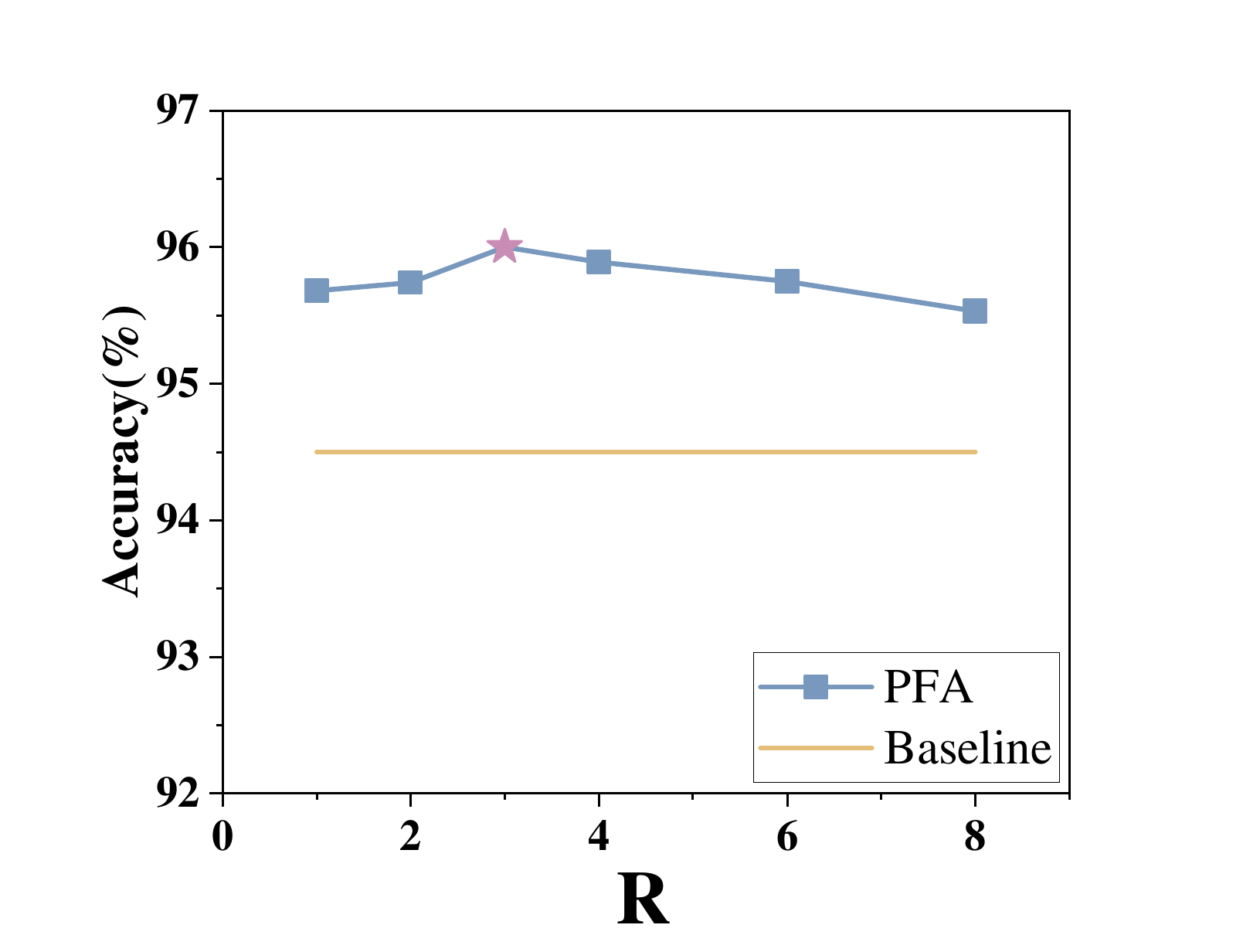}}
        \centerline{CIFAR10}
    \end{minipage}
    \begin{minipage}{0.48\linewidth}
		\centerline{\includegraphics[scale=0.19]{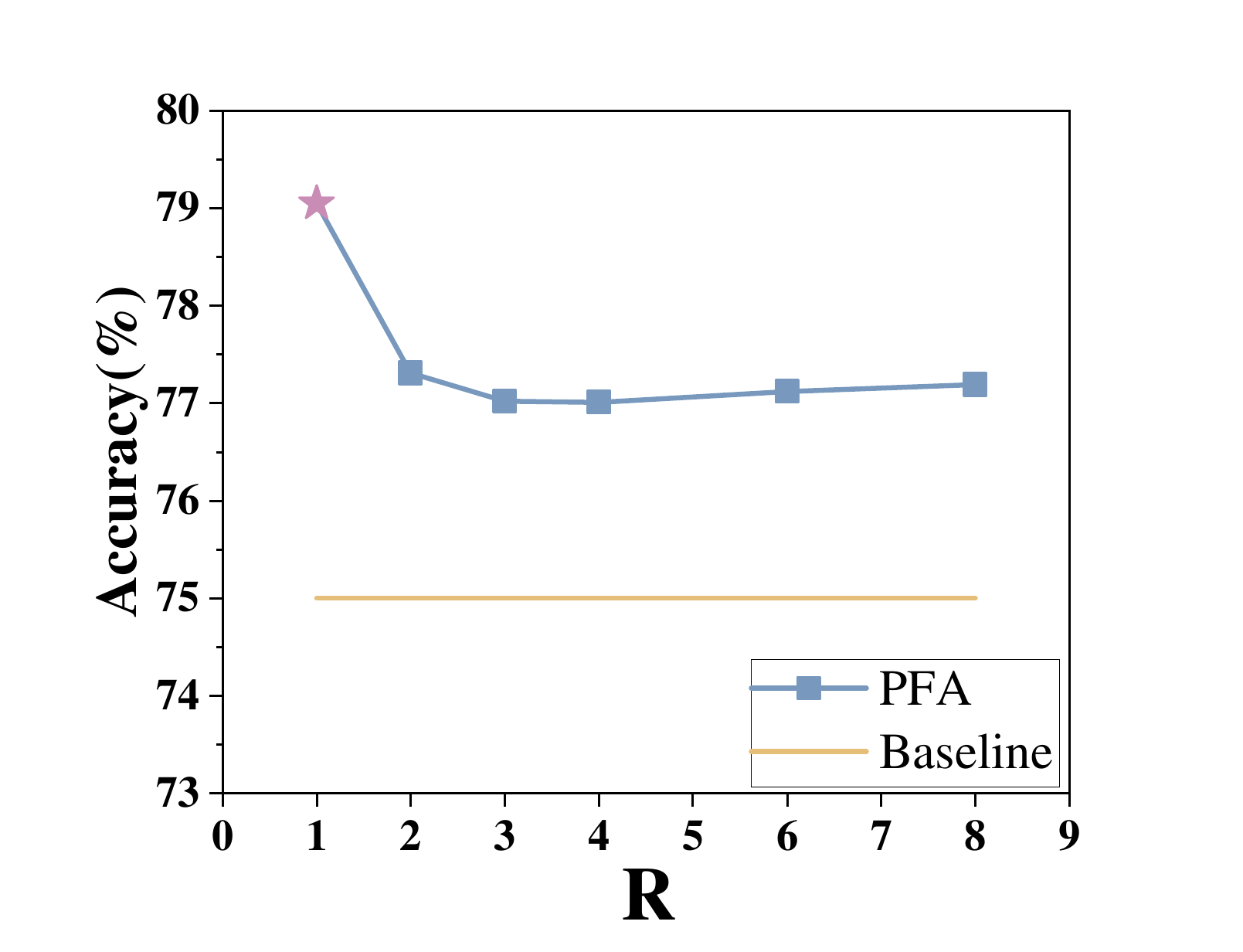}}
		\centerline{CIFAR100}
    \end{minipage}
        \begin{minipage}{0.48\linewidth}
		\centerline{\includegraphics[scale=0.19]{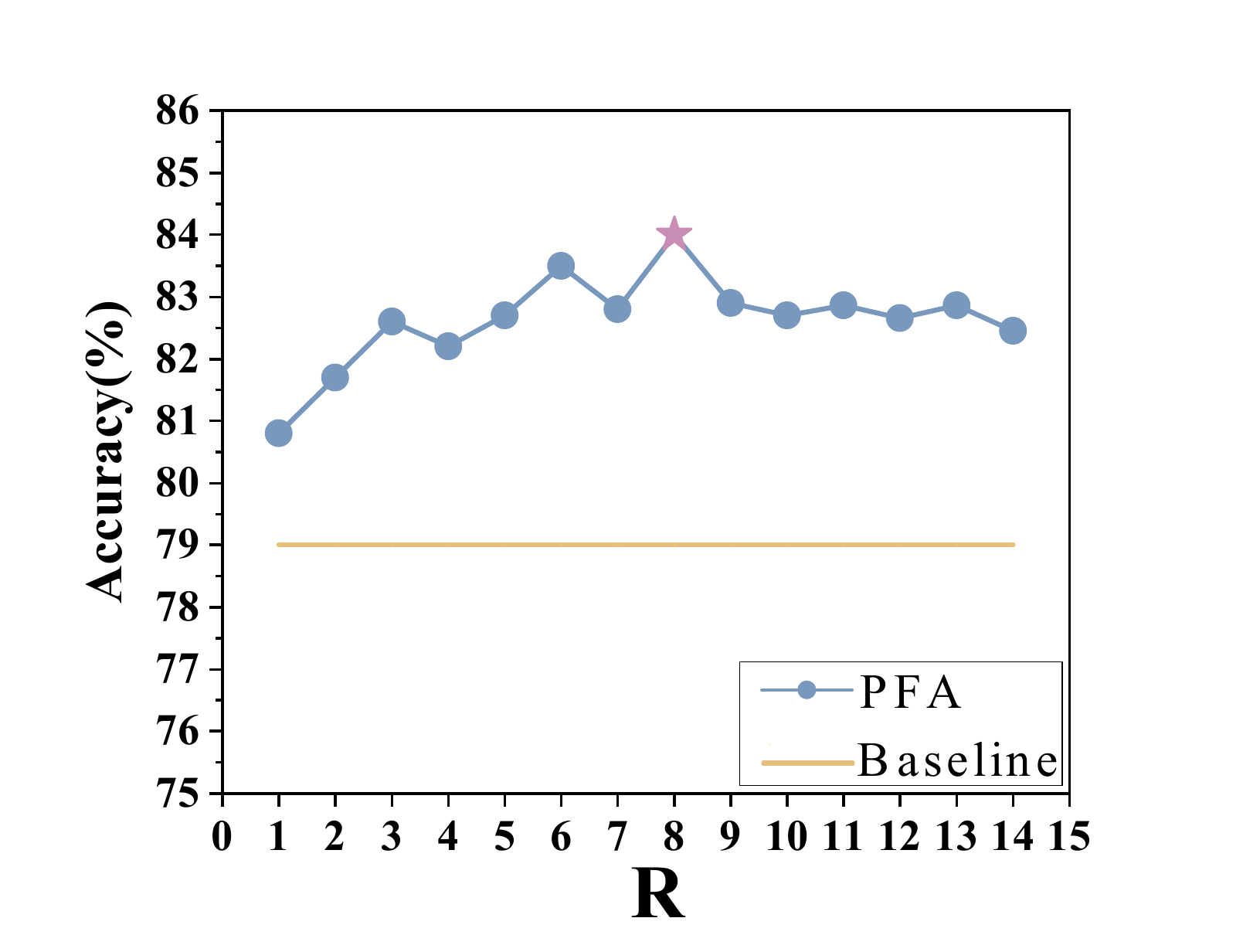}}
		\centerline{CIFAR10DVS}
    \end{minipage}
    \begin{minipage}{0.48\linewidth}
        \centerline{\includegraphics[scale=0.19]{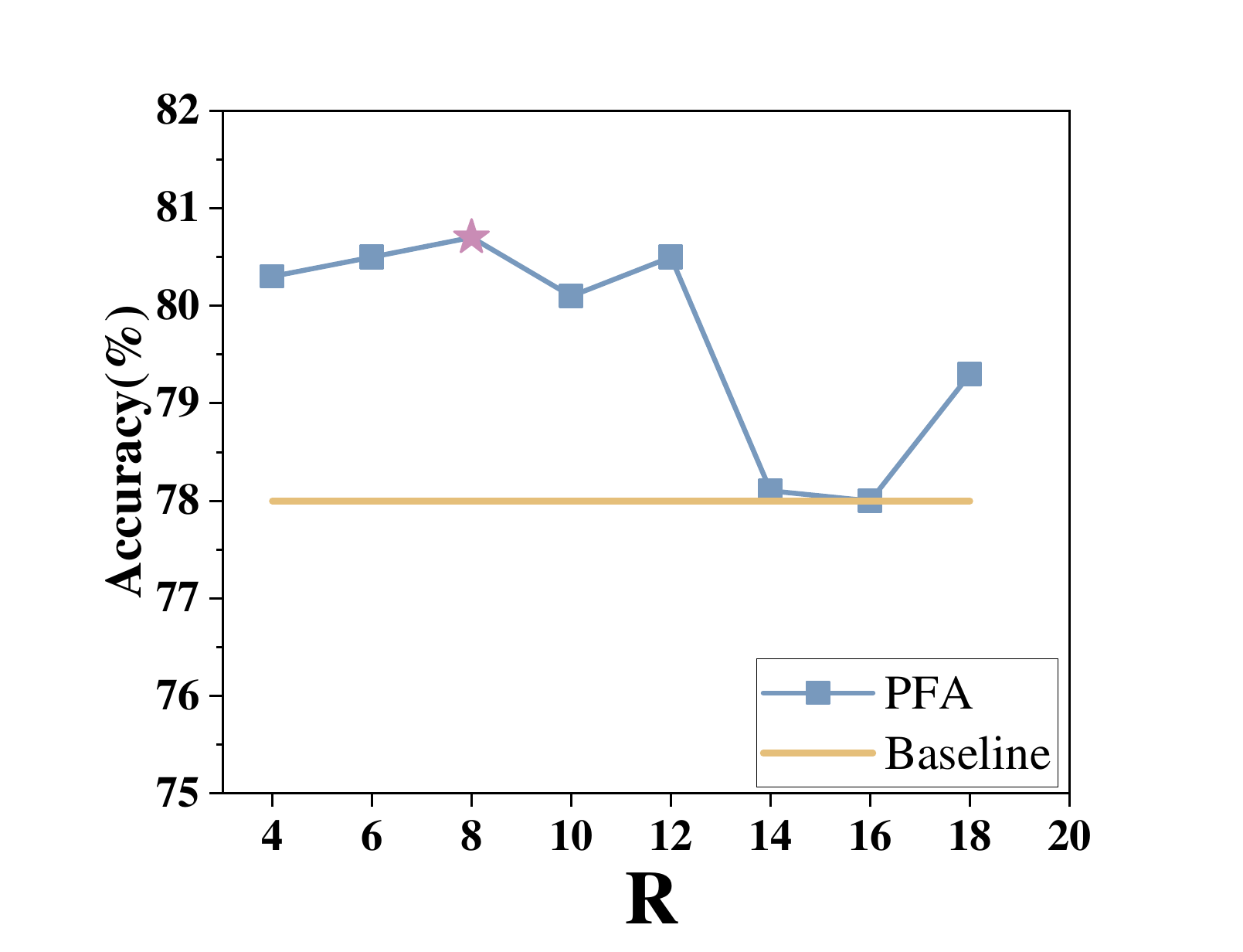}}
        \centerline{NCaltech101}
    \end{minipage}
    \caption{Accuracy on four tested datasets with different $R$. PFA achieves best performance when $r=3,1,8,8$ correspondingly.}
    \vspace{-0.5cm}
    \label{fig:rank}
\end{figure}
Figure \ref{fig:rank} demonstrates the performance with different $R$. The outcome reveals that for dynamic datasets, the best rank option falls around $\frac{T}{2}$ while for static datasets, the best value of $R$ (also known as connecting factor) is very small. 

Dynamic datasets exhibit significant variations in content across different time steps. Conversely, static datasets involve the duplication of frames, where each frame is replicated $T$ times as input to the network. Consequently, this duplication results in the attention map being subjected to a low-rank condition.
Considering a comprehensive assessment of both experimental outcomes and theoretical insights, our conclusion asserts that the precision of the attention map in faithfully mirroring the input tensor is unnecessary. Such fidelity does not inherently facilitate the effective concentration on pivotal components. Furthermore, our findings unveil an intriguing phenomenon: even in scenarios where SNNs exhibit considerable sparsity, the rank of the attention map can remain markedly low. This revelation implies the presence of untapped optimization potential within the realm of SNN sparsity.
\par

{It is} worth noting that on the CIFAR10-DVS dataset, when $R=1$, the performance of PFA is comparable to TCJA (see Table \ref{tabresult} and does not show significant advantages. This phenomenon strongly confirms our previous statement that TCJA is a "rank-1" attention, a specal case of PFA, highlighting the necessity and advantages of non-"rank-1" attention modules.

\subsection{Ablation Study}

\begin{figure}
    \centering
   \includegraphics[width=0.8\linewidth]{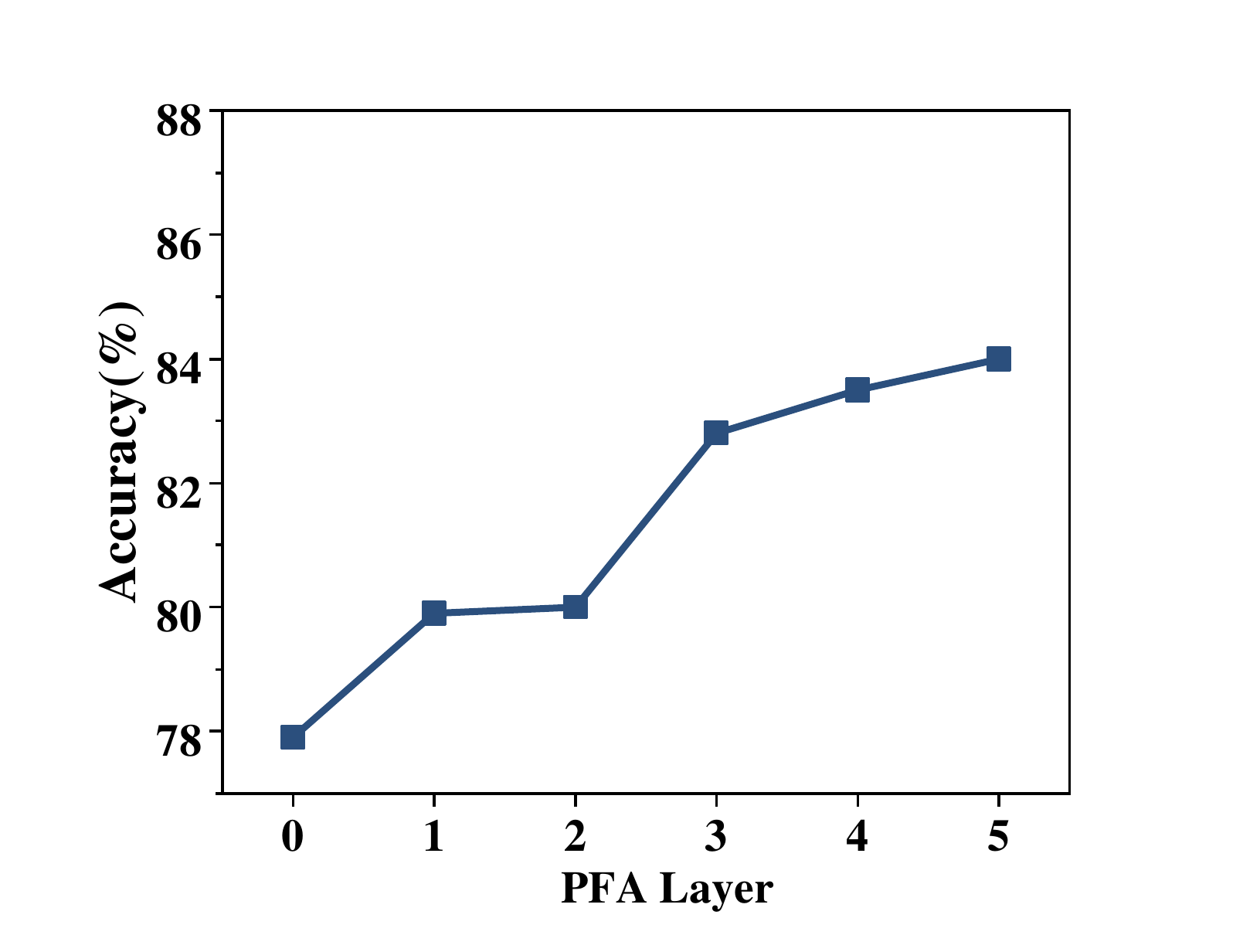}
    \caption{{Ablation study on PFA layers. The x-axis represents the amount of PFA inserted. The accuracy improves with the insertion of PFA layer.}}
    \label{fig:ablation_layers}
    \vspace{-0.5cm}
\end{figure}

\begin{figure}
    \centering
    \includegraphics[width=0.9\linewidth]{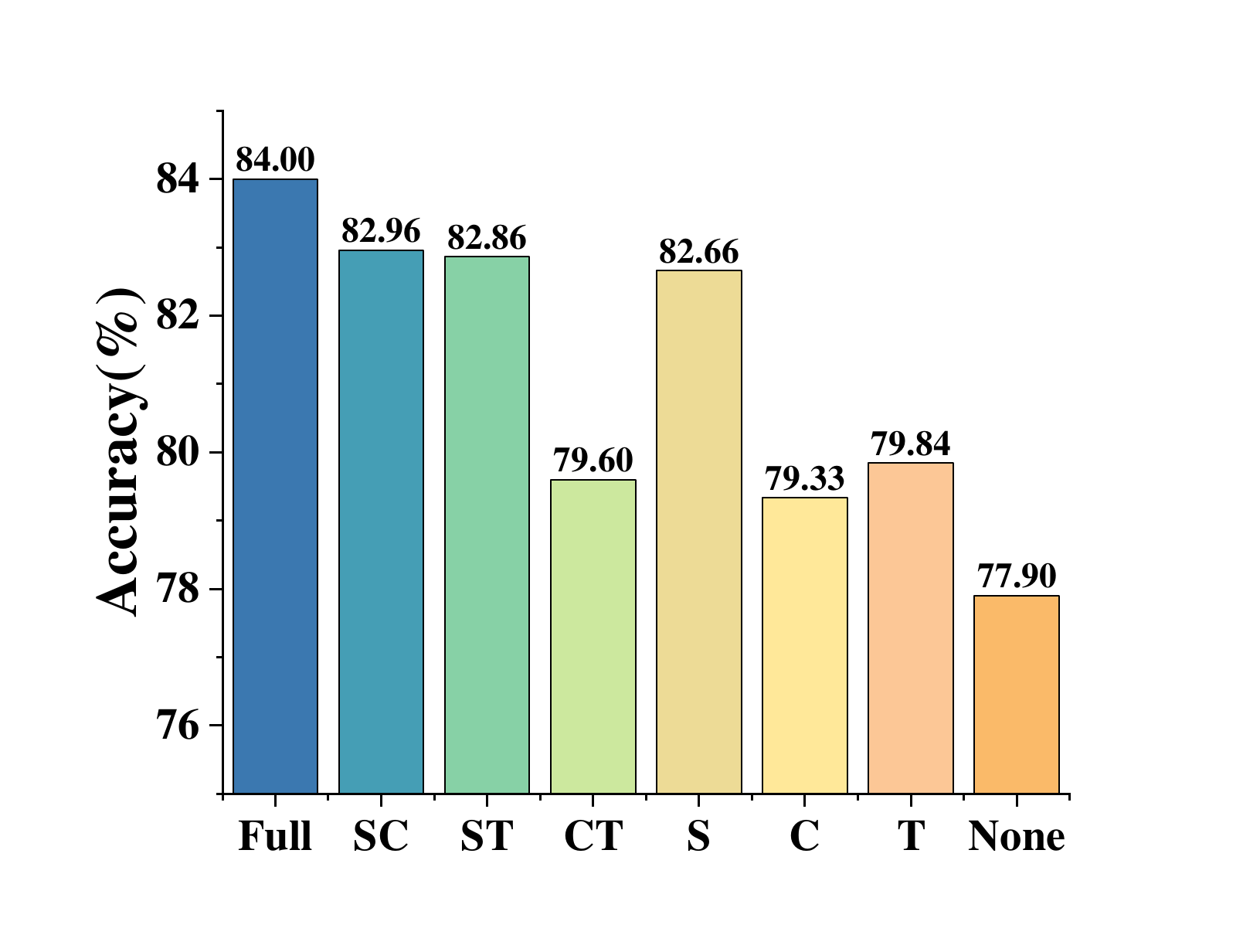}
    \caption{{Ablation study on attention dimensions. T/C/S denotes temporal/channel/spatial correspondingly. SC denotes spatial-channel joint attention and so on. Full means paying attention to three dimensions at the same time, \emph{i.e.} a standard PFA module.}}
    \label{fig:ablation}
    \vspace{-0.5cm}
\end{figure}
\begin{figure*}[h]
    \centering
    \includegraphics[scale=0.17]{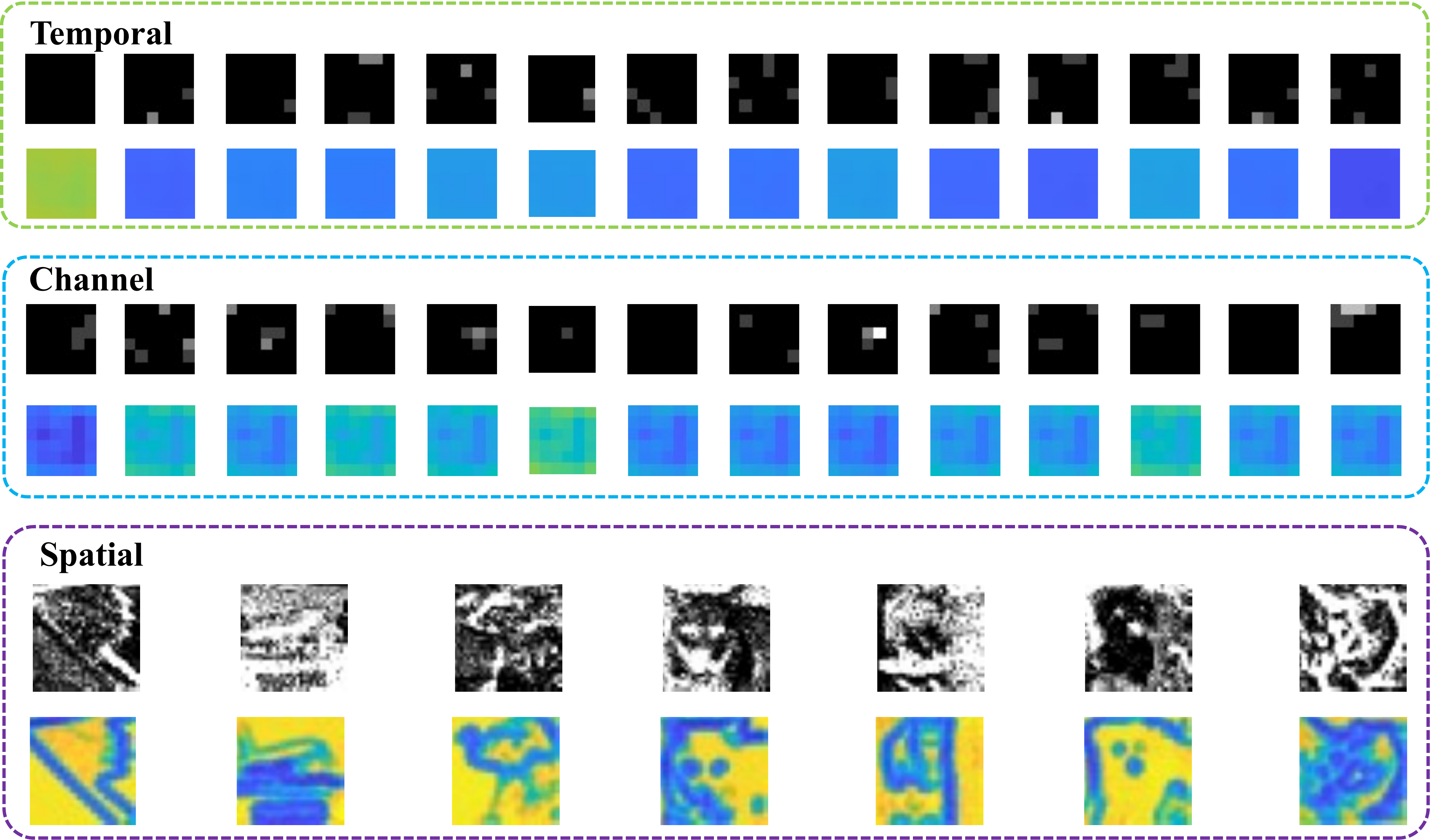}
    \caption{Attention visualization on CIFAR10DVS dataset. Temporal and channel parts are taken from the deep layer of the VGG network. 
    The spatial layer visualizations are extracted from the shallow layer of the network due to the inherent complexity of the information contained in the deeper layers, which may impede human comprehension and hinder the interpretability of our analysis.}
    \label{fig:vatt}
\end{figure*}
In this section, we perform ablation experiments from both global and local perspectives to validate the efficacy of PFA on CIFAR10DVS \citep{cifar10} Initially, we vary the number of added PFA layers in the network to assess the impact on classification accuracy. Subsequently, we set the three projections obtained within the PFA to constant matrices, with each element assigned a value of 1, aiming to eliminate attention {to} specific dimensions. This helps explore the influence of different attention dimensions on classification results. Finally, for a more comprehensive insight into the influence of PFA on the training process, we conduct ablation experiments specifically involving the insertion of PFA.
\par
We incorporate varying numbers of PFAs into the VGG network and present the results in Figure \ref{fig:ablation_layers}. In this experiment, the value of $R$ is held constant at 8. {It is evident that the network’s accuracy exhibits a noticeable improvement with more PFA layers.} Remarkably, even the addition of a single layer leads to enhanced accuracy in the network.
\par
In order to explore how attention in temporal-channel-spatial dimensions affects classification accuracy, we conduct an ablation by fixing the corresponding projection matrix obtained by LPST to an all-one tensor. As shown in Figure \ref{fig:ablation}, whichever dimension is ablated will adversely affect the results. Ablating temporal and ablating channels have similar effects. This means that PFA treats temporal and channel dimensions equally. However, when we ablate the spatial dimension, the accuracy of the network drops significantly (79.60\%) but still outperforms the baseline (77.90\%). Whichever dimension we ablate, the performance drops. {When we restrict the focus dimensions to only one, namely, spatial, temporal, or channel, we observe {another} decrease in the network's performance compared to when attention is distributed across two dimensions. The retention of spatial attention significantly impacts accuracy, followed by temporal attention.} This observation highlights that PFA effectively directs attention to distinct regions across multiple dimensions, aligning with the intended purpose of attention modules. Specifically, PFA exhibits attention distribution across all three dimensions, albeit with varying weights. Notably, the spatial dimension emerges as the most significant, followed by the time and channel dimensions, in terms of attention importance.

\subsection{Attention Visualization\label{section:vatten}}

In order to visually represent the attention distribution of the PFA module, we have generated an attention map in PFA and depicted it in Figure \ref{fig:vatt}. The attention maps corresponding to the temporal and channel dimensions in the figure are obtained from the deep layer of the network. This choice was made because deep layers tend to provide a more accurate reflection of the attention's role. {Conversely, the spatial attention map is obtained from the network's input layer, where the shallow graphic features are more distinct and conspicuous, facilitating analysis and observation.} Figure \ref{fig:vatt} also illustrates the distinctive attention distributions in the temporal, channel, and spatial dimensions achieved by PFA. This finding aligns with the conclusions drawn from our ablation study, further validating our earlier assertions.

\section{Conclusions}
{In this paper, we introduce a novel attention module named PFA, which combines learning and the generation of a temporal-channel-spatial attention map through CP decomposition.} Specifically, we propose the LPST to learn a projection of the input tensor that can extract essential information for AMC to compose the attention map. In the AMC, we utilize the reverse process of CP decomposition to generate the attention map. After analyzing the rank of the raw input tensor and the best option for connecting factor $R$, we propose a principle to search for the best $R$. Remarkably, our findings suggest that even a low rank can lead to outstanding performance. On classification tasks, regardless of the form of the dataset, dynamic or static, PFA achieves the SOTA performance. {When we extend our evaluation of PFA to image generation tasks, it still shows its capacity to enhance performance on this task as well. Furthermore, additional ablation studies bolster the rationale behind PFA's efficacy, while visualizing the attention distribution of PFA offers intuitive insights into its substantial impact.}
\par
In the future, we plan to extend the applicability of the PFA model to various demanding intelligent applications and deploy the model on a neuromorphic chip.

\bibliographystyle{plainnat}
\bibliography{cas-refs}

\begin{thebibliography}{70}
\providecommand{\natexlab}[1]{#1}
\providecommand{\url}[1]{\texttt{#1}}
\expandafter\ifx\csname urlstyle\endcsname\relax
  \providecommand{\doi}[1]{doi: #1}\else
  \providecommand{\doi}{doi: \begingroup \urlstyle{rm}\Url}\fi

\bibitem[Acar et~al.(2011)Acar, Dunlavy, Kolda, and M{\o}rup]{scalable}
Evrim Acar, Daniel~M. Dunlavy, Tamara~G. Kolda, and Morten M{\o}rup.
\newblock Scalable tensor factorizations for incomplete data.
\newblock \emph{Chemometrics and Intelligent Laboratory Systems}, 106\penalty0 (1):\penalty0 41--56, March 2011.
\newblock \doi{10.1016/j.chemolab.2010.08.004}.
\newblock URL \url{https://doi.org/10.1016/j.chemolab.2010.08.004}.

\bibitem[Babiloni et~al.(2023)Babiloni, Tanay, Deng, Maggioni, and Zafeiriou]{babiloni2023factorized}
Francesca Babiloni, Thomas Tanay, Jiankang Deng, Matteo Maggioni, and Stefanos Zafeiriou.
\newblock Factorized dynamic fully-connected layers for neural networks.
\newblock In \emph{Proceedings of the IEEE/CVF International Conference on Computer Vision}, pages 1374--1383, 2023.

\bibitem[Bohte et~al.(2002)Bohte, Kok, and Poutr{\'{e}}]{BP}
Sander~M. Bohte, Joost~N. Kok, and Han~La Poutr{\'{e}}.
\newblock Error-backpropagation in temporally encoded networks of spiking neurons.
\newblock \emph{Neurocomputing}, 48\penalty0 (1-4):\penalty0 17--37, October 2002.
\newblock \doi{10.1016/s0925-2312(01)00658-0}.
\newblock URL \url{https://doi.org/10.1016/s0925-2312(01)00658-0}.

\bibitem[Cai et~al.(2022)Cai, Sun, Liu, Cui, Wang, Xia, Yao, and Guo]{TCS_Cai}
Wuque Cai, Hongze Sun, Rui Liu, Yan Cui, Jun Wang, Yang Xia, Dezhong Yao, and Daqing Guo.
\newblock A spatial-channel-temporal-fused attention for spiking neural networks, 2022.
\newblock URL \url{https://arxiv.org/abs/2209.10837}.

\bibitem[Chen et~al.(2020)Chen, Zhu, Sun, He, Li, Shen, and Yu]{chenwanli}
Wanli Chen, Xinge Zhu, Ruoqi Sun, Junjun He, Ruiyu Li, Xiaoyong Shen, and Bei Yu.
\newblock Tensor low-rank reconstruction for semantic segmentation.
\newblock In Andrea Vedaldi, Horst Bischof, Thomas Brox, and Jan-Michael Frahm, editors, \emph{Computer Vision -- ECCV 2020}, pages 52--69, Cham, 2020. Springer International Publishing.
\newblock ISBN 978-3-030-58520-4.

\bibitem[Cheng et~al.(2021)Cheng, Hao, Xu, and Xu]{LISNN}
Xiang Cheng, Yunzhe Hao, Jiaming Xu, and Bo~Xu.
\newblock Lisnn: improving spiking neural networks with lateral interactions for robust object recognition.
\newblock In \emph{Proceedings of the Twenty-Ninth International Joint Conference on Artificial Intelligence}, IJCAI'20, 2021.
\newblock ISBN 9780999241165.

\bibitem[Deng(2012)]{minist}
Li~Deng.
\newblock The mnist database of handwritten digit images for machine learning research [best of the web].
\newblock \emph{IEEE Signal Processing Magazine}, 29\penalty0 (6):\penalty0 141--142, 2012.
\newblock \doi{10.1109/MSP.2012.2211477}.

\bibitem[Deng et~al.(2021)Deng, Li, Zhang, and Gu]{deng2021temporal}
Shikuang Deng, Yuhang Li, Shanghang Zhang, and Shi Gu.
\newblock {Temporal Efficient Training of Spiking Neural Network via Gradient Re-weighting}.
\newblock In \emph{International Conference on Learning Representations (ICLR)}, 2021.

\bibitem[Diehl et~al.(2015)Diehl, Neil, Binas, Cook, Liu, and Pfeiffer]{Conversion2}
Peter~U. Diehl, Daniel Neil, Jonathan Binas, Matthew Cook, Shih-Chii Liu, and Michael Pfeiffer.
\newblock Fast-classifying, high-accuracy spiking deep networks through weight and threshold balancing.
\newblock In \emph{2015 International Joint Conference on Neural Networks (IJCNN)}, pages 1--8, 2015.
\newblock \doi{10.1109/IJCNN.2015.7280696}.

\bibitem[Fang et~al.(2020)Fang, Chen, Ding, Chen, Yu, Zhou, Tian, and other contributors]{SpikingJelly}
Wei Fang, Yanqi Chen, Jianhao Ding, Ding Chen, Zhaofei Yu, Huihui Zhou, Yonghong Tian, and other contributors.
\newblock Spikingjelly.
\newblock \url{https://github.com/fangwei123456/spikingjelly}, 2020.
\newblock Accessed: 2023-01-16.

\bibitem[Fang et~al.(2021{\natexlab{a}})Fang, Yu, Chen, Huang, Masquelier, and Tian]{fang2021deep}
Wei Fang, Zhaofei Yu, Yanqi Chen, Tiejun Huang, Timoth{\'e}e Masquelier, and Yonghong Tian.
\newblock Deep residual learning in spiking neural networks.
\newblock \emph{Advances in Neural Information Processing Systems}, 34:\penalty0 21056--21069, 2021{\natexlab{a}}.

\bibitem[Fang et~al.(2021{\natexlab{b}})Fang, Yu, Chen, Masquelier, Huang, and Tian]{fang2021incorporating}
Wei Fang, Zhaofei Yu, Yanqi Chen, Timoth\'ee Masquelier, Tiejun Huang, and Yonghong Tian.
\newblock {Incorporating Learnable Membrane Time Constant To Enhance Learning of Spiking Neural Networks}.
\newblock In \emph{Proceedings of the IEEE/CVF International Conference on Computer Vision (ICCV)}, pages 2661--2671, 2021{\natexlab{b}}.

\bibitem[H{\aa}stad(1990)]{np_hard1}
Johan H{\aa}stad.
\newblock Tensor rank is {NP}-complete.
\newblock \emph{Journal of Algorithms}, 11\penalty0 (4):\penalty0 644--654, December 1990.
\newblock \doi{10.1016/0196-6774(90)90014-6}.
\newblock URL \url{https://doi.org/10.1016/0196-6774(90)90014-6}.

\bibitem[He et~al.(2016)He, Zhang, Ren, and Sun]{ResNet}
Kaiming He, Xiangyu Zhang, Shaoqing Ren, and Jian Sun.
\newblock Deep residual learning for image recognition.
\newblock In \emph{2016 {IEEE} Conference on Computer Vision and Pattern Recognition ({CVPR})}. {IEEE}, June 2016.
\newblock \doi{10.1109/cvpr.2016.90}.
\newblock URL \url{https://doi.org/10.1109/cvpr.2016.90}.

\bibitem[Hodgkin and Huxley(1952)]{HH}
A.~L. Hodgkin and A.~F. Huxley.
\newblock A quantitative description of membrane current and its application to conduction and excitation in nerve.
\newblock \emph{The Journal of Physiology}, 117\penalty0 (4):\penalty0 500--544, August 1952.
\newblock \doi{10.1113/jphysiol.1952.sp004764}.
\newblock URL \url{https://doi.org/10.1113/jphysiol.1952.sp004764}.

\bibitem[Hu et~al.(2018{\natexlab{a}})Hu, Shen, and Sun]{SENet}
Jie Hu, Li~Shen, and Gang Sun.
\newblock Squeeze-and-excitation networks.
\newblock In \emph{Proceedings of the IEEE conference on computer vision and pattern recognition}, pages 7132--7141, 2018{\natexlab{a}}.

\bibitem[Hu et~al.(2018{\natexlab{b}})Hu, Tang, and Pan]{SDRN}
Yangfan Hu, Huajin Tang, and Gang Pan.
\newblock Spiking deep residual network, 2018{\natexlab{b}}.
\newblock URL \url{https://arxiv.org/abs/1805.01352}.

\bibitem[Hu et~al.(2022)Hu, Wu, Deng, and Li]{ADRN}
Yifan Hu, Yujie Wu, Lei Deng, and Guoqi Li.
\newblock Advancing deep residual learning by solving the crux of degradation in spiking neural networks, 2022.
\newblock URL \url{https://arxiv.org/abs/2201.07209}.

\bibitem[Izhikevich(2003)]{izhikevich2003simple}
Eugene~M Izhikevich.
\newblock Simple model of spiking neurons.
\newblock \emph{IEEE Transactions on neural networks}, 14\penalty0 (6):\penalty0 1569--1572, 2003.

\bibitem[Janzamin et~al.(2019)Janzamin, Ge, Kossaifi, and Anandkumar]{8918128}
Majid Janzamin, Rong Ge, Jean Kossaifi, and Anima Anandkumar.
\newblock 2019.

\bibitem[Kamata et~al.(2022)Kamata, Mukuta, and Harada]{fsvae}
Hiromichi Kamata, Yusuke Mukuta, and Tatsuya Harada.
\newblock Fully spiking variational autoencoder.
\newblock In \emph{Proceedings of the AAAI Conference on Artificial Intelligence}, volume~36, pages 7059--7067, 2022.

\bibitem[Kim and Panda(2021)]{kim2021optimizing}
Youngeun Kim and Priyadarshini Panda.
\newblock {Optimizing Deeper Spiking Neural Networks for Dynamic Vision Sensing}.
\newblock \emph{Neural Networks}, 144:\penalty0 686--698, 2021.

\bibitem[Kim et~al.(2022)Kim, Li, Park, Venkatesha, and Panda]{kim2022neural}
Youngeun Kim, Yuhang Li, Hyoungseob Park, Yeshwanth Venkatesha, and Priyadarshini Panda.
\newblock Neural architecture search for spiking neural networks.
\newblock In \emph{Computer Vision--ECCV 2022: 17th European Conference, Tel Aviv, Israel, October 23--27, 2022, Proceedings, Part XXIV}, pages 36--56. Springer, 2022.

\bibitem[Kolda and Bader(2009{\natexlab{a}})]{CPdeconstruction}
Tamara~G. Kolda and Brett~W. Bader.
\newblock Tensor decompositions and applications.
\newblock \emph{{SIAM} Review}, 51\penalty0 (3):\penalty0 455--500, August 2009{\natexlab{a}}.
\newblock \doi{10.1137/07070111x}.
\newblock URL \url{https://doi.org/10.1137/07070111x}.

\bibitem[Kolda and Bader(2009{\natexlab{b}})]{kolda}
Tamara~G. Kolda and Brett~W. Bader.
\newblock Tensor decompositions and applications.
\newblock \emph{SIAM Review}, 51\penalty0 (3):\penalty0 455--500, 2009{\natexlab{b}}.
\newblock \doi{10.1137/07070111X}.
\newblock URL \url{https://doi.org/10.1137/07070111X}.

\bibitem[Kossaifi et~al.(2017)Kossaifi, Lipton, Kolbeinsson, Khanna, Furlanello, and Anandkumar]{kossaifi2017}
Jean Kossaifi, Zachary~C. Lipton, Arinbjorn Kolbeinsson, Aran Khanna, Tommaso Furlanello, and Anima Anandkumar.
\newblock Tensor regression networks, 2017.
\newblock URL \url{https://arxiv.org/abs/1707.08308}.

\bibitem[Kossaifi et~al.(2019)Kossaifi, Toisoul, Bulat, Panagakis, Hospedales, and Pantic]{Kossaifi2019FactorizedHC}
Jean Kossaifi, Antoine Toisoul, Adrian Bulat, Yannis Panagakis, Timothy~M. Hospedales, and Maja Pantic.
\newblock Factorized higher-order cnns with an application to spatio-temporal emotion estimation.
\newblock \emph{2020 IEEE/CVF Conference on Computer Vision and Pattern Recognition (CVPR)}, pages 6059--6068, 2019.
\newblock URL \url{https://api.semanticscholar.org/CorpusID:214743123}.

\bibitem[Kossaifi et~al.(2020)Kossaifi, Toisoul, Bulat, Panagakis, Hospedales, and Pantic]{Kossaifi2020}
Jean Kossaifi, Antoine Toisoul, Adrian Bulat, Yannis Panagakis, Timothy~M. Hospedales, and Maja Pantic.
\newblock Factorized higher-order cnns with an application to spatio-temporal emotion estimation.
\newblock In \emph{2020 IEEE/CVF Conference on Computer Vision and Pattern Recognition (CVPR)}. IEEE, June 2020.
\newblock \doi{10.1109/cvpr42600.2020.00610}.
\newblock URL \url{http://dx.doi.org/10.1109/CVPR42600.2020.00610}.

\bibitem[Krizhevsky et~al.()Krizhevsky, Nair, and Hinton]{cifar10}
Alex Krizhevsky, Vinod Nair, and Geoffrey Hinton.
\newblock Cifar-10 (canadian institute for advanced research).
\newblock URL \url{http://www.cs.toronto.edu/~kriz/cifar.html}.

\bibitem[Krizhevsky et~al.(2009)Krizhevsky, Hinton, et~al.]{krizhevsky2009learning}
Alex Krizhevsky, Geoffrey Hinton, et~al.
\newblock Learning multiple layers of features from tiny images.
\newblock 2009.

\bibitem[Lapique(1907)]{LIF}
Louis Lapique.
\newblock Recherches quantitatives sur l'excitation electrique des nerfs traitee comme une polarization.
\newblock \emph{Journal of Physiology and Pathology}, 9:\penalty0 620--635, 1907.

\bibitem[Lau et~al.(2024)Lau, Po, and Rehman]{Lau2024}
Kin~Wai Lau, Lai-Man Po, and Yasar Abbas~Ur Rehman.
\newblock Large separable kernel attention: Rethinking the large kernel attention design in cnn.
\newblock \emph{Expert Systems with Applications}, 236:\penalty0 121352, February 2024.
\newblock ISSN 0957-4174.
\newblock \doi{10.1016/j.eswa.2023.121352}.
\newblock URL \url{http://dx.doi.org/10.1016/j.eswa.2023.121352}.

\bibitem[Li et~al.(2017)Li, Liu, Ji, Li, and Shi]{CIFAR10DVS}
Hongmin Li, Hanchao Liu, Xiangyang Ji, Guoqi Li, and Luping Shi.
\newblock {CIFAR}10-{DVS}: An event-stream dataset for object classification.
\newblock \emph{Frontiers in Neuroscience}, 11, May 2017.
\newblock \doi{10.3389/fnins.2017.00309}.
\newblock URL \url{https://doi.org/10.3389/fnins.2017.00309}.

\bibitem[Li et~al.(2021)Li, Guo, Zhang, Deng, Hai, and Gu]{li2021differentiable}
Yuhang Li, Yufei Guo, Shanghang Zhang, Shikuang Deng, Yongqing Hai, and Shi Gu.
\newblock {Differentiable Spike: Rethinking Gradient-Descent for Training Spiking Neural Networks}.
\newblock In \emph{Advances in Neural Information Processing Systems (NeurIPS)}, volume~34, pages 23426--23439, 2021.

\bibitem[Liu and Parhi(2023)]{liu2023tensor}
Xingyi Liu and Keshab~K Parhi.
\newblock Tensor decomposition for model reduction in neural networks: A review [feature].
\newblock \emph{IEEE Circuits and Systems Magazine}, 23\penalty0 (2):\penalty0 8--28, 2023.

\bibitem[Maass(1997)]{MAASS19971659}
Wolfgang Maass.
\newblock Networks of spiking neurons: The third generation of neural network models.
\newblock \emph{Neural Networks}, 10\penalty0 (9):\penalty0 1659--1671, 1997.
\newblock ISSN 0893-6080.
\newblock \doi{https://doi.org/10.1016/S0893-6080(97)00011-7}.
\newblock URL \url{https://www.sciencedirect.com/science/article/pii/S0893608097000117}.

\bibitem[Meng et~al.(2022)Meng, Xiao, Yan, Wang, Lin, and Luo]{meng2022training}
Qingyan Meng, Mingqing Xiao, Shen Yan, Yisen Wang, Zhouchen Lin, and Zhi-Quan Luo.
\newblock {Training High-Performance Low-Latency Spiking Neural Networks by Differentiation on Spike Representation}.
\newblock \emph{ArXiv preprint arXiv:2205.00459}, 2022.

\bibitem[Mnih et~al.(2014)Mnih, Heess, Graves, and Kavukcuoglu]{Google_Attention_2014}
Volodymyr Mnih, Nicolas Heess, Alex Graves, and Koray Kavukcuoglu.
\newblock Recurrent models of visual attention, 2014.
\newblock URL \url{https://arxiv.org/abs/1406.6247}.

\bibitem[Novikov et~al.(2015)Novikov, Podoprikhin, Osokin, and Vetrov]{Novikov2015TensorizingNN}
Alexander Novikov, Dmitry Podoprikhin, Anton Osokin, and Dmitry~P. Vetrov.
\newblock Tensorizing neural networks.
\newblock In \emph{Neural Information Processing Systems}, 2015.
\newblock URL \url{https://api.semanticscholar.org/CorpusID:290242}.

\bibitem[Orchard et~al.(2015)Orchard, Jayawant, Cohen, and Thakor]{NCALTECH101}
Garrick Orchard, Ajinkya Jayawant, Gregory~K. Cohen, and Nitish Thakor.
\newblock Converting static image datasets to spiking neuromorphic datasets using saccades.
\newblock \emph{Frontiers in Neuroscience}, 9, November 2015.
\newblock \doi{10.3389/fnins.2015.00437}.
\newblock URL \url{https://doi.org/10.3389/fnins.2015.00437}.

\bibitem[Qiu et~al.(2023)Qiu, Wang, Luan, Zhu, Wu, Zhang, and Deng]{vtsnn}
Xue-Rui Qiu, Zhao-Rui Wang, Zheng Luan, Rui-Jie Zhu, Xiao Wu, Ma-Lu Zhang, and Liang-Jian Deng.
\newblock Vtsnn: A virtual temporal spiking neural network.
\newblock \emph{Frontiers in neuroscience}, 17:\penalty0 1091097, 2023.

\bibitem[Qiu et~al.(2024)Qiu, Zhu, Chou, Wang, Deng, and Li]{gated}
Xuerui Qiu, Rui-Jie Zhu, Yuhong Chou, Zhaorui Wang, Liang-jian Deng, and Guoqi Li.
\newblock Gated attention coding for training high-performance and efficient spiking neural networks.
\newblock In \emph{Proceedings of the AAAI Conference on Artificial Intelligence}, volume~38, pages 601--610, 2024.

\bibitem[Rathi and Roy(2021)]{rathi2021diet}
Nitin Rathi and Kaushik Roy.
\newblock Diet-snn: A low-latency spiking neural network with direct input encoding and leakage and threshold optimization.
\newblock \emph{IEEE Transactions on Neural Networks and Learning Systems (TNNLS)}, 2021.

\bibitem[Rathi et~al.()Rathi, Srinivasan, Panda, and Roy]{rathienabling}
Nitin Rathi, Gopalakrishnan Srinivasan, Priyadarshini Panda, and Kaushik Roy.
\newblock Enabling deep spiking neural networks with hybrid conversion and spike timing dependent backpropagation.
\newblock In \emph{International Conference on Learning Representations (ICLR)}.

\bibitem[Rueckauer et~al.(2017)Rueckauer, Lungu, Hu, Pfeiffer, and Liu]{Conversion1}
Bodo Rueckauer, Iulia-Alexandra Lungu, Yuhuang Hu, Michael Pfeiffer, and Shih-Chii Liu.
\newblock Conversion of continuous-valued deep networks to efficient event-driven networks for image classification.
\newblock \emph{Frontiers in Neuroscience}, 11, December 2017.
\newblock \doi{10.3389/fnins.2017.00682}.
\newblock URL \url{https://doi.org/10.3389/fnins.2017.00682}.

\bibitem[Sidiropoulos et~al.(2017)Sidiropoulos, De~Lathauwer, Fu, Huang, Papalexakis, and Faloutsos]{7891546}
Nicholas~D. Sidiropoulos, Lieven De~Lathauwer, Xiao Fu, Kejun Huang, Evangelos~E. Papalexakis, and Christos Faloutsos.
\newblock Tensor decomposition for signal processing and machine learning.
\newblock \emph{IEEE Transactions on Signal Processing}, 65\penalty0 (13):\penalty0 3551--3582, 2017.
\newblock \doi{10.1109/TSP.2017.2690524}.

\bibitem[Simonyan and Zisserman(2015)]{VGG}
Karen Simonyan and Andrew Zisserman.
\newblock Very deep convolutional networks for large-scale image recognition.
\newblock \emph{CoRR}, abs/1409.1556, 2015.

\bibitem[Wang et~al.(2023)Wang, Liu, Cui, and Han]{wang2023inertial}
Qingsong Wang, Zehui Liu, Chunfeng Cui, and Deren Han.
\newblock Inertial accelerated sgd algorithms for solving large-scale lower-rank tensor cp decomposition problems.
\newblock \emph{Journal of Computational and Applied Mathematics}, 423:\penalty0 114948, 2023.

\bibitem[Wang et~al.(2022)Wang, Zhang, Chen, and Qu]{wang2022signed}
Yuchen Wang, Malu Zhang, Yi~Chen, and Hong Qu.
\newblock Signed neuron with memory: Towards simple, accurate and high-efficient ann-snn conversion.
\newblock In \emph{International Joint Conference on Artificial Intelligence}, 2022.

\bibitem[Wei et~al.(2023)Wei, Zhang, Qu, Belatreche, Zhang, and Chen]{wei2023temporal}
Wenjie Wei, Malu Zhang, Hong Qu, Ammar Belatreche, Jian Zhang, and Hong Chen.
\newblock Temporal-coded spiking neural networks with dynamic firing threshold: Learning with event-driven backpropagation.
\newblock In \emph{Proceedings of the IEEE/CVF International Conference on Computer Vision}, pages 10552--10562, 2023.

\bibitem[Wei et~al.(2024)Wei, Zhang, Zhang, Belatreche, Wu, Xu, Qiu, Chen, Yang, and Li]{Wenjie2024}
Wenjie Wei, Malu Zhang, Jilin Zhang, Ammar Belatreche, Jibin Wu, Zijing Xu, Xuerui Qiu, Hong Chen, Yang Yang, and Haizhou Li.
\newblock Event-driven learning for spiking neural networks.
\newblock \emph{arXiv preprint arXiv:2403.00270}, 2024.

\bibitem[Wu et~al.(2021{\natexlab{a}})Wu, Chua, Zhang, Li, Li, and Tan]{wu2021tandem}
Jibin Wu, Yansong Chua, Malu Zhang, Guoqi Li, Haizhou Li, and Kay~Chen Tan.
\newblock A tandem learning rule for effective training and rapid inference of deep spiking neural networks.
\newblock \emph{IEEE Transactions on Neural Networks and Learning Systems}, 2021{\natexlab{a}}.

\bibitem[Wu et~al.(2021{\natexlab{b}})Wu, Xu, Han, Zhou, Zhang, Li, and Tan]{wu2021progressive}
Jibin Wu, Chenglin Xu, Xiao Han, Daquan Zhou, Malu Zhang, Haizhou Li, and Kay~Chen Tan.
\newblock Progressive tandem learning for pattern recognition with deep spiking neural networks.
\newblock \emph{IEEE Transactions on Pattern Analysis and Machine Intelligence}, 44\penalty0 (11):\penalty0 7824--7840, 2021{\natexlab{b}}.

\bibitem[Wu et~al.(2018)Wu, Deng, Li, Zhu, and Shi]{STBP}
Yujie Wu, Lei Deng, Guoqi Li, Jun Zhu, and Luping Shi.
\newblock Spatio-temporal backpropagation for training high-performance spiking neural networks.
\newblock \emph{Frontiers in Neuroscience}, 12, May 2018.
\newblock \doi{10.3389/fnins.2018.00331}.
\newblock URL \url{https://doi.org/10.3389/fnins.2018.00331}.

\bibitem[Wu et~al.(2019{\natexlab{a}})Wu, Deng, Li, Zhu, Xie, and Shi]{Wu2019}
Yujie Wu, Lei Deng, Guoqi Li, Jun Zhu, Yuan Xie, and Luping Shi.
\newblock Direct training for spiking neural networks: Faster, larger, better.
\newblock \emph{Proceedings of the {AAAI} Conference on Artificial Intelligence}, 33:\penalty0 1311--1318, July 2019{\natexlab{a}}.
\newblock \doi{10.1609/aaai.v33i01.33011311}.
\newblock URL \url{https://doi.org/10.1609/aaai.v33i01.33011311}.

\bibitem[Wu et~al.(2019{\natexlab{b}})Wu, Deng, Li, Zhu, Xie, and Shi]{wu2019direct}
Yujie Wu, Lei Deng, Guoqi Li, Jun Zhu, Yuan Xie, and Luping Shi.
\newblock {Direct Training for Spiking Neural Networks: Faster, Larger, Better}.
\newblock In \emph{Proceedings of the AAAI Conference on Artificial Intelligence (AAAI)}, pages 1311--1318, 2019{\natexlab{b}}.
\newblock \doi{10.1609/aaai.v33i01.33011311}.

\bibitem[Wu et~al.(2021{\natexlab{c}})Wu, Zhang, Lin, Li, Wang, and Tang]{wu2021liaf}
Zhenzhi Wu, Hehui Zhang, Yihan Lin, Guoqi Li, Meng Wang, and Ye~Tang.
\newblock {LIAF-Net: Leaky Integrate and Analog Fire Network for Lightweight and Efficient Spatiotemporal Information Processing}.
\newblock \emph{IEEE Transactions on Neural Networks and Learning Systems}, pages 1--14, 2021{\natexlab{c}}.
\newblock \doi{10.1109/TNNLS.2021.3073016}.

\bibitem[Wu et~al.(2022)Wu, Huang, Deng, Dou, and Meng]{NEURIPS2022_acbfe708}
Zhong-Cheng Wu, Ting-Zhu Huang, Liang-Jian Deng, Hong-Xia Dou, and Deyu Meng.
\newblock Tensor wheel decomposition and its tensor completion application.
\newblock In S.~Koyejo, S.~Mohamed, A.~Agarwal, D.~Belgrave, K.~Cho, and A.~Oh, editors, \emph{Advances in Neural Information Processing Systems}, volume~35, pages 27008--27020. Curran Associates, Inc., 2022.
\newblock URL \url{https://proceedings.neurips.cc/paper_files/paper/2022/file/acbfe708197ff78ad04cc1beb1710979-Paper-Conference.pdf}.

\bibitem[Xiao et~al.(2017)Xiao, Rasul, and Vollgraf]{FashionMNISTAN}
Han Xiao, Kashif Rasul, and Roland Vollgraf.
\newblock Fashion-mnist: a novel image dataset for benchmarking machine learning algorithms.
\newblock \emph{ArXiv}, abs/1708.07747, 2017.
\newblock URL \url{https://api.semanticscholar.org/CorpusID:702279}.

\bibitem[Xu et~al.(2023)Xu, Lv, Chu, and Li]{xu2023}
Zhihao Xu, Zhiqiang Lv, Benjia Chu, and Jianbo Li.
\newblock Fast autoregressive tensor decomposition for online real-time traffic flow prediction.
\newblock \emph{Knowledge-Based Systems}, 282:\penalty0 111125, December 2023.
\newblock ISSN 0950-7051.
\newblock \doi{10.1016/j.knosys.2023.111125}.
\newblock URL \url{http://dx.doi.org/10.1016/j.knosys.2023.111125}.

\bibitem[Yang and Hospedales(2016)]{Yang2016DeepMR}
Yongxin Yang and Timothy~M. Hospedales.
\newblock Deep multi-task representation learning: A tensor factorisation approach.
\newblock \emph{ArXiv}, abs/1605.06391, 2016.
\newblock URL \url{https://api.semanticscholar.org/CorpusID:3047732}.

\bibitem[Yao et~al.(2021)Yao, Gao, Zhao, Wang, Lin, Yang, and Li]{TASNN}
Man Yao, Huanhuan Gao, Guangshe Zhao, Dingheng Wang, Yihan Lin, Zhaoxu Yang, and Guoqi Li.
\newblock Temporal-wise attention spiking neural networks for event streams classification.
\newblock In \emph{2021 {IEEE}/{CVF} International Conference on Computer Vision ({ICCV})}. {IEEE}, October 2021.
\newblock \doi{10.1109/iccv48922.2021.01006}.
\newblock URL \url{https://doi.org/10.1109/iccv48922.2021.01006}.

\bibitem[Yao et~al.(2023)Yao, Zhao, Zhang, Hu, Deng, Tian, Xu, and Li]{yao2023attention}
Man Yao, Guangshe Zhao, Hengyu Zhang, Yifan Hu, Lei Deng, Yonghong Tian, Bo~Xu, and Guoqi Li.
\newblock Attention spiking neural networks.
\newblock \emph{IEEE Transactions on Pattern Analysis and Machine Intelligence}, 2023.

\bibitem[Zhan et~al.(2023)Zhan, Liu, Xie, Zhang, and Sun]{Zhan2023}
Qiugang Zhan, Guisong Liu, Xiurui Xie, Malu Zhang, and Guolin Sun.
\newblock Bio-inspired active learning method in spiking neural network.
\newblock \emph{Knowledge-Based Systems}, 261:\penalty0 110193, February 2023.
\newblock ISSN 0950-7051.
\newblock \doi{10.1016/j.knosys.2022.110193}.
\newblock URL \url{http://dx.doi.org/10.1016/j.knosys.2022.110193}.

\bibitem[Zhang et~al.(2021)Zhang, Wang, Wu, Belatreche, Amornpaisannon, Zhang, Miriyala, Qu, Chua, Carlson, et~al.]{zhang2021rectified}
Malu Zhang, Jiadong Wang, Jibin Wu, Ammar Belatreche, Burin Amornpaisannon, Zhixuan Zhang, Venkata Pavan~Kumar Miriyala, Hong Qu, Yansong Chua, Trevor~E Carlson, et~al.
\newblock Rectified linear postsynaptic potential function for backpropagation in deep spiking neural networks.
\newblock \emph{IEEE transactions on neural networks and learning systems}, 33\penalty0 (5):\penalty0 1947--1958, 2021.

\bibitem[Zhang and Li(2019)]{Spike-Train}
Wenrui Zhang and Peng Li.
\newblock \emph{Spike-train level backpropagation for training deep recurrent spiking neural networks}.
\newblock Curran Associates Inc., Red Hook, NY, USA, 2019.

\bibitem[Zheng et~al.(2020)Zheng, Wu, Deng, Hu, and Li]{DTBN}
Hanle Zheng, Yujie Wu, Lei Deng, Yifan Hu, and Guoqi Li.
\newblock Going deeper with directly-trained larger spiking neural networks, 2020.
\newblock URL \url{https://arxiv.org/abs/2011.05280}.

\bibitem[Zheng et~al.(2021)Zheng, Wu, Deng, Hu, and Li]{zheng2021going}
Hanle Zheng, Yujie Wu, Lei Deng, Yifan Hu, and Guoqi Li.
\newblock {Going Deeper With Directly-Trained Larger Spiking Neural Networks}.
\newblock In \emph{Proceedings of the AAAI Conference on Artificial Intelligence (AAAI)}, pages 11062--11070, 2021.

\bibitem[Zhou et~al.(2023)Zhou, Zhu, He, Wang, YAN, Tian, and Yuan]{zhou2023spikformer}
Zhaokun Zhou, Yuesheng Zhu, Chao He, Yaowei Wang, Shuicheng YAN, Yonghong Tian, and Li~Yuan.
\newblock Spikformer: When spiking neural network meets transformer.
\newblock In \emph{The Eleventh International Conference on Learning Representations}, 2023.
\newblock URL \url{https://openreview.net/forum?id=frE4fUwz_h}.

\bibitem[Zhu et~al.(2022)Zhu, Zhao, Zhang, Deng, Duan, Zhang, and Deng]{TCJA}
Rui-Jie Zhu, Qihang Zhao, Tianjing Zhang, Haoyu Deng, Yule Duan, Malu Zhang, and Liang-Jian Deng.
\newblock Tcja-snn: Temporal-channel joint attention for spiking neural networks, 2022.
\newblock URL \url{https://arxiv.org/abs/2206.10177}.

\end{thebibliography}

\end{document}